\documentclass[10pt,twocolumn,letterpaper]{article}

\usepackage{cvpr}              % To produce the CAMERA-READY version
\newcommand{\ours}{\textsc{VeGAS}}

\newcommand{\vp}[1]{}
\newcommand{\sj}[1]{}
\newcommand{\cb}[1]{}
\newcommand{\anja}[1]{}
\newcommand{\hritik}[1]{}

\usepackage{microtype}

\usepackage{amsmath}
\usepackage{amssymb}
\usepackage{mathtools}
\usepackage{amsthm}
\usepackage{dsfont}
\usepackage{nicefrac}
\usepackage{bm}

\allowdisplaybreaks

\usepackage{graphicx}
\usepackage{subcaption}
\usepackage{sidecap}
\usepackage{caption}
\usepackage{wrapfig}
\usepackage{float}
\usepackage{placeins}

\usepackage{booktabs}
\usepackage{colortbl}
\usepackage{xcolor}
\usepackage{tabularx}
\usepackage{multirow}

\usepackage{listings}
\usepackage{mdframed}
\usepackage[skins,theorems,breakable]{tcolorbox}

\tcbset{
  aibox/.style={
    width=\linewidth,
    top=7pt,
    bottom=2pt,
    colback=blue!6!white,
    colframe=black,
    colbacktitle=black,
    enhanced,
    center,
    attach boxed title to top left={yshift=-0.1in,xshift=0.15in},
    boxed title style={boxrule=0pt,colframe=white,},
  }
}
\newtcolorbox{AIbox}[2][]{aibox,title=#2,#1}
\newtcolorbox{examplebox}[1]{
    enhanced,
    drop shadow=black!10!white,
    left=4mm,
    right=4mm,
    top=2mm,
    bottom=2mm,
    boxsep=0mm,
    rounded corners,
    title=#1,
    fontupper=\footnotesize\linespread{0.9}\fontfamily{lmr}\selectfont,}

\usepackage{enumitem}
\usepackage{soul}
\usepackage{url}
\usepackage{textcomp}
\usepackage{pifont}
\usepackage{fontawesome5}

\definecolor{listinggray}{gray}{0.9}
\definecolor{headerred}{RGB}{178, 34, 34}
\definecolor{lightblue}{rgb}{0.22,0.45,0.70}
\definecolor{darkblue}{rgb}{0, 0, 0.5}
\colorlet{oursrow}{blue!8}
\definecolor{baselinerow}{RGB}{230,230,230}

\definecolor{cvprblue}{rgb}{0.21,0.49,0.74}
\usepackage[pagebackref,breaklinks,colorlinks,allcolors=cvprblue]{hyperref}

\def\paperID{*****} % *** Enter the Paper ID here
\def\confName{CVPR}
\def\confYear{2026}

\title{Think Twice, Act Once: Verifier-Guided Action Selection For Embodied Agents}

\author{Nishad Singhi \quad Christian Bialas \quad Snehal Jauhri \quad Vignesh Prasad \\
Georgia Chalvatzaki \quad Marcus Rohrbach \quad Anna Rohrbach \\[6pt]
Technical University of Darmstadt \& hessian.AI \\
{\tt nishad.singhi@tu-darmstadt.de} \\[2pt]
{ \href{https://nishadsinghi.github.io/vegas}{\faGlobe~Webpage} \hspace{2cm} \href{https://github.com/nishadsinghi/vegas}{\faGithub~Code}
}
}

\begin{document}
\maketitle
\begin{abstract}
    Building generalist embodied agents capable of solving complex real-world tasks remains a fundamental challenge in AI. Multimodal Large Language Models (MLLMs) have significantly advanced the reasoning capabilities of such agents through strong vision-language knowledge and chain-of-thought (CoT) reasoning, yet remain brittle when faced with challenging out-of-distribution scenarios.
    To address this, we propose \textbf{Verifier-Guided Action Selection (\ours{})}, a test-time framework designed to improve the robustness of MLLM-based embodied agents through an explicit verification step. At inference time, rather than committing to a single decoded action, \ours{} samples an ensemble of candidate actions and uses a generative verifier to identify the most reliable choice, without modifying the underlying policy. Crucially, we find that using an MLLM off-the-shelf as a verifier yields no improvement, motivating our LLM-driven data synthesis strategy, which automatically constructs a diverse curriculum of failure cases to expose the verifier to a rich distribution of potential errors at training time. Across embodied reasoning benchmarks spanning the Habitat and ALFRED environments, \ours{} consistently improves generalization, achieving up to a \textbf{36\% relative performance gain} over strong CoT baselines on the most challenging multi-object, long-horizon tasks. %Code is available at \url{https://github.com/nishadsinghi/vegas}.
\end{abstract}

\section{Introduction}

\begin{figure}[t]
    \centering
    \includegraphics[width=\linewidth]{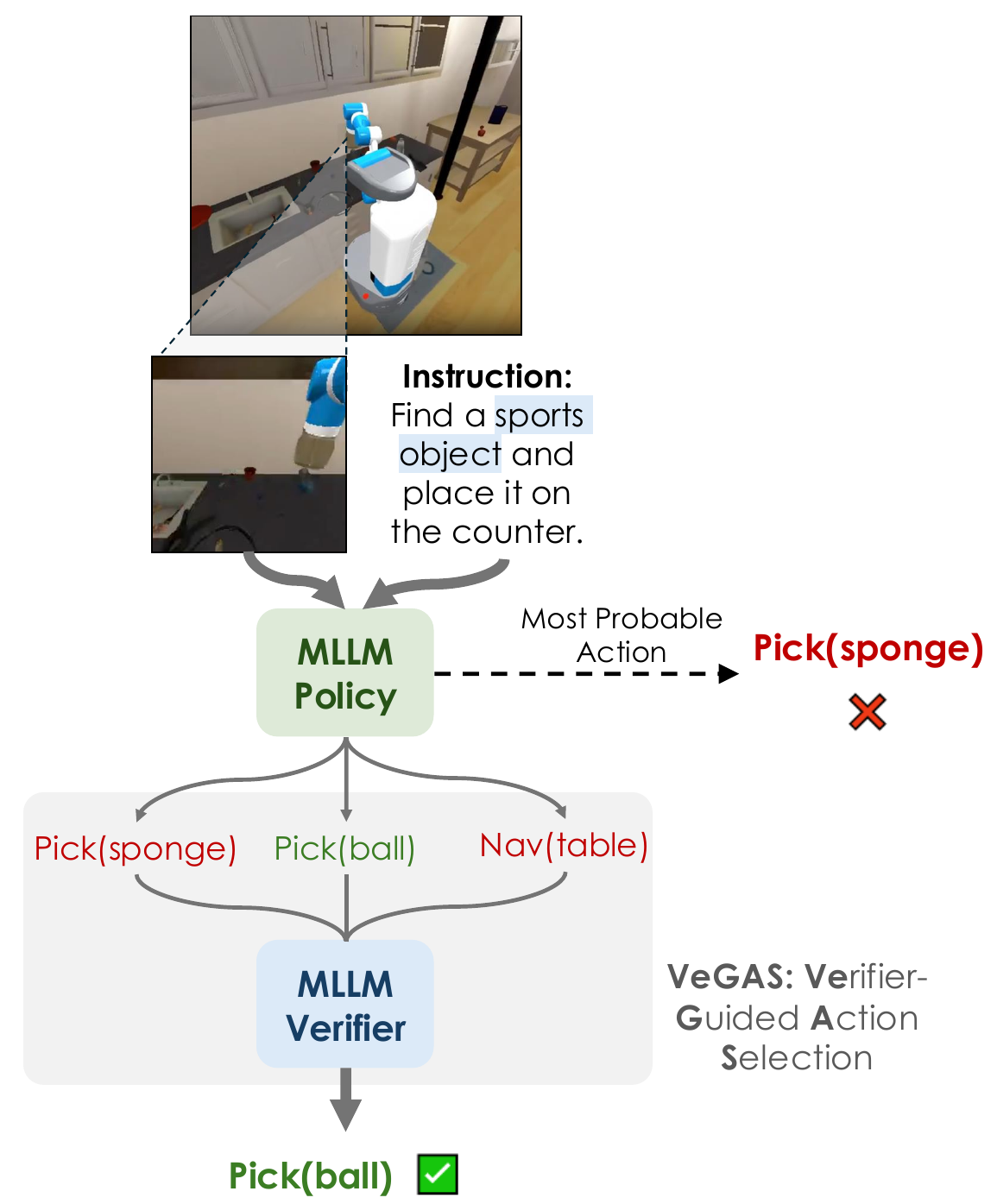}
    \caption{\textbf{Overview of Verifier-Guided Action Selection (\ours{}).} Given a task instruction (e.g., ``Find a sports object and place it on the counter''), standard policies decode a single action that may be incorrect under distribution shifts (right). \ours{}, instead, samples multiple candidate actions with reasoning traces, evaluates them using a generative verifier, and executes only the highest-scored action (bottom). This test-time verification strategy substantially improves robustness in challenging out-of-distribution scenarios involving long-horizon tasks.}
    \label{fig:teaser}
\end{figure}

A longstanding goal in AI is to create embodied agents that operate autonomously in physical environments to accomplish complex tasks specified through natural language~\citep{ding2023task,li2024llm}, such as navigating to target locations~\citep{qiao2025open} and manipulating everyday objects~\citep{song2023llm,singh2023progprompt}. Recently, Multimodal Large Language Models (MLLMs), pretrained on Internet-scale vision-language data, have emerged as a promising foundation for building such agents, owing to their strong perceptual and linguistic generalization~\citep{yang2025embodiedbench,huang2022language,szot2023large}. While early efforts relied on the zero-shot capabilities of MLLMs~\citep{huang2022language,sarch2023open,ahn2022can}, finetuning on embodied data---either via supervised learning~\citep{yang2024embodied} or reinforcement learning~\citep{szot2023large,szot2024grounding,zhai2024fine}---yields significant improvements. More recently, incorporating Chain-of-Thought (CoT) reasoning has further enhanced decision-making by enabling agents to reason step-by-step before acting~\citep{wei2022chain,mu2023embodiedgpt,zawalski2024robotic,zhai2024fine}. Despite this progress, MLLM-based embodied agents remain brittle in out-of-distribution scenarios and long-horizon settings~\citep{yang2025embodiedbench}. For instance, an agent might reliably execute ``bring me a banana'' but fail when the same goal is phrased as ``bring me a yellow curved fruit.'' Similarly, an agent trained on single-object pick-and-place may fail on a multi-step task such as cleaning an apple and placing it in a cabinet.

We observe that a key factor underlying these failures is that agents cannot recognize mistakes in their reasoning process and correct them at test time. In particular, they commit to a single greedily decoded action at each step with no opportunity for self-correction. In contrast, humans routinely consider multiple candidate actions, mentally evaluate their likely outcomes, and commit only to the most promising one, effectively performing verification before acting. This idea has a direct computational analogue: recent work on scaling test-time compute shows that sampling multiple candidate solutions and selecting the best one via a learned verifier substantially improves LLM performance in domains such as coding and mathematics~\citep{cobbe2021training,brown2024large,snell2024scaling}. However, extending verification to high-level embodied reasoning poses distinct challenges: unlike in mathematics or code, embodied agents operate under partial observability and must reason about semantic task progression from egocentric observations alone, with compounding errors in long-horizon plans. Yet verification for high-level embodied reasoning remains largely unexplored.

To bridge this gap, we introduce \textbf{Verifier-Guided Action Selection (\ours{})}, a framework that improves the robustness of MLLM-based embodied agents by incorporating an explicit verification step at test time. Concretely, at each timestep \ours{} samples multiple candidate actions from the policy, each accompanied by a Chain-of-Thought rationale. A learned generative verifier~\citep{zhang2025generative,ankner2024critique} then evaluates each candidate by producing an explicit reasoning trace followed by a correctness judgement, and the agent executes only the highest-scoring action (see Figure~\ref{fig:teaser} for an overview). A critical finding is that using an MLLM as a verifier off-the-shelf yields no improvement over the base policy — general-purpose language understanding alone is insufficient for embodied verification. This motivates specialized verifier training; however, standard embodied datasets contain only successful demonstrations, providing no signal for what constitutes an incorrect action. To address this, we introduce an LLM-driven pipeline that automatically synthesizes diverse, realistic failure trajectories paired with verification annotations, constructing a rich curriculum of both correct and incorrect examples without additional human data collection.

\ours{} yields consistent improvements across embodied reasoning benchmarks in Habitat 2.0~\citep{szot2021habitat,szot2023large} and AI2-THOR~\citep{shridhar2020alfred}, raising average success rates from 65\% to 71\% on LangR and from 44\% to 49\% on EB-ALFRED over strong CoT baselines, while also improving significantly larger off-the-shelf policies.

\noindent Our key contributions are:
\begin{enumerate}[leftmargin=*,itemsep=2pt,topsep=4pt]
    \item We propose Verifier-Guided Action Selection (\ours{}), a test-time verification framework for high-level embodied reasoning that samples diverse candidate actions and uses a learned generative verifier to select the most reliable one at each timestep. We find that using MLLMs off-the-shelf as verifiers does not improve performance, motivating our specialized training pipeline.
    % \item We introduce an automated, LLM-driven pipeline that synthesizes diverse and realistic failure trajectories paired with verification annotations, addressing the challenge that standard embodied datasets contain only successful demonstrations and thus provide no signal for verifier training.
    \item Training a verifier requires demonstrations of both correct and incorrect actions, yet embodied datasets typically lack the latter. To address this, we introduce an automated, LLM-driven pipeline that synthesizes diverse and realistic failure trajectories paired with verification annotations, without additional human data collection.
    \item Extensive experiments in Habitat 2.0~\citep{szot2021habitat,szot2023large} and AI2-THOR~\citep{shridhar2020alfred} show that \ours{} consistently improves over strong CoT baselines, scales more effectively with test-time compute than Self-Consistency~\citep{wang2022self}, and generalizes to large, off-the-shelf policies.
\end{enumerate}
\section{Related Work}

% \ns{TODO}

\noindent\textbf{Foundation Models for Embodied Agents.}
Multimodal LLMs have proven useful for developing intelligent systems that perceive and interact with an environment~\citep{du2024embspatial,yang2025embodiedbench,li2024embodied}. They have shown strong generalization skills in areas such as Language-guided Navigation~\citep{qiao2025open,dorbala2023can,schumann2024velma,lin2025navcot}, Task Planning~\citep{song2023llm,singh2023progprompt,sun2023adaplanner,huang2022language}, and Embodied Decision Making~\citep{li2022pre,sarch2023open,szot2023large,yan2025efficient,du2023guiding}.
While such approaches have mostly used off-the-shelf or fine-tuned LLMs/VLMs,
recent works show that Chain-of-Thought (CoT)~\citep{wei2022chain} can further improve performance via multi-modal reasoning~\citep{mu2023embodiedgpt,zawalski2024robotic}, sub-goal consistency~\citep{lu2025thinkbot}, or spatial reasoning~\citep{sun2025emma}. Some works have explored improving multi-agent embodied cooperation through coordinated planning~\citep{liu2024capo} and tree-search-based collaborative deliberation~\citep{zu2025collaborative}; while these methods orchestrate multiple off-the-shelf LLM agents via structured communication and joint plan search, our work targets single-agent action reliability through a dedicated verifier trained on synthetically generated failure data.
%In this work, we explore enhancing the CoT reasoning for embodied tasks via incorporating generative verifiers~\citep{zhang2025generative}, which have proven useful for improving LLM reasoning, as discussed next.

\noindent\textbf{Verifiers.} Verification has recently emerged as a key strategy for improving LLM reasoning by evaluating and selecting among candidate solutions. Early work trained separate verifiers to score solutions between 0 and 1, selecting the solution with the highest score as the final answer (i.e., Best-of-N)~\citep{cobbe2021training, yu2024ovm}. Recent work has shown the advantages of generative verifiers that produce verification rationales (i.e., critiques/corrections), consistently outperforming discriminative verifiers while also enhancing explainability~\citep{zhang2025generative,ankner2024critique,singhi2025solve}. In multimodal settings, vision-language reward models extend verification to visual outcomes~\citep{zhang2025generative,sun2025mm}, and discriminative verifiers have been applied to low-level control via VLA models~\citep{kwok2025robomonkey}. In contrast, our work is the first to apply generative verifiers to high-level embodied reasoning, with an emphasis on challenging scenarios requiring novel behaviors and robustness to linguistic variations. Concurrently,~\citep{hong2026learning} propose generating and scoring multiple candidate actions along with test-time training for embodied agents.

\noindent\textbf{Embodied Agent Benchmarks.} Several simulation platforms support embodied AI research, including AI2-THOR~\citep{kolve2017ai2,procthor,robothor,manipulathor} and Habitat~\citep{habitat2019iccv,szot2021habitat,puig2024habitat}, with benchmarks spanning diverse task complexities~\citep{shridhar2020alfred,padmakumar2022teach,kant2022housekeep,li2023behavior,shridhar2021alfworld}.
LangR~\citep{szot2023large}, built on Habitat 2.0~\citep{szot2021habitat}, evaluates out-of-distribution generalization through two axes: \textit{paraphrastic robustness} (e.g., ``pick up a banana'' $\rightarrow$ ``pick up a yellow curved fruit'') and \textit{behavioral generalization} (e.g., extending single-object tasks to multi-object variants).
ALFRED~\citep{shridhar2020alfred} contains 25K language-annotated household tasks with both high-level goals and low-level instructions across six core task types including pick-and-place, clean-and-place, and examine-in-light scenarios. TEACH~\citep{padmakumar2022teach} extends this with over 3,000 human-human dialogues for interactive task completion ranging from ``Make Coffee'' to ``Prepare Breakfast''.
More recently, EmbodiedBench~\citep{yang2025embodiedbench} offers a comprehensive evaluation framework with 1,128 tasks across hierarchical action levels, from high-level planning to low-level motor control, assessing capabilities such as spatial awareness and long-horizon planning.

\section{Preliminaries}
\label{sec:background}

We now formalize the embodied decision-making setup and the policy architecture that serves as the foundation for our approach.

\textbf{Problem Formulation.}
We formulate the agent's task as a sequential decision-making problem under partial observability. The agent's objective is to generate a sequence of actions $a_1, \ldots, a_T$ to accomplish a goal specified as a natural language instruction $I$ (e.g., ``Bring an item that can be used for cutting to the left counter''). At each timestep $t$, the agent receives an egocentric RGB image $o_t$ as its observation. The agent's true underlying state $s_t$ is not directly accessible. The agent must decide on its next action $a_t$ based on the goal $I$ and its history $h_t$ composed of all its past observations and actions $(o_1, a_1, ..., o_{t-1}, a_{t-1}, o_t)$. Our aim is to learn a policy $\pi$ that maps the goal and history to the next action: $\pi(a_t | I, o_{1:t}, a_{1:t-1})$. The action space $A$ consists of high-level semantic actions, such as \texttt{pick(apple)} and \texttt{navigate(table)}. Following prior work~\citep{szot2023large, yang2025embodiedbench}, we assume an oracle low-level policy that executes these high-level actions once selected.

\textbf{Policy Architecture.}
We instantiate the policy $\pi$ as a multimodal large language model (MLLM) that takes visual and text tokens as input and autoregressively generates text tokens as output.
Given the goal in the form of a text instruction $I$ and the history $h_t$ as input, the policy autoregressively emits an output token sequence
$y_t = (c_t, a_t)$. Here, $c_t$ is an optional chain-of-thought rationale (a possibly empty sequence of text tokens) followed
by the action token sequence $a_t$. Following~\citet{szot2024grounding}, the actions are encoded in natural language (e.g., ``\texttt{pick(apple)}''), and the output can be extracted to obtain the action $a_t$, which is sent to the environment to be executed by the low-level policy~\citep{szot2023large, yang2025embodiedbench}.

\textbf{Policy Training.}
We train the policy via imitation learning on expert demonstrations $\mathcal{D} = \{\tau\}$, where each trajectory $\tau = \big(I, (o_1, a_1), \ldots, (o_T, a_T)\big)$ depicts a successful execution of the task. The model is fine-tuned via supervised next-token prediction to maximize the likelihood of the expert output $y_t = (c_t, a_t)$, computing the loss only over output tokens, including the CoT prefix $c_t$ when present.

\section{\ours{}: Verifier-Guided Action Selection}

\begin{figure}[t]
    \centering
    \includegraphics[width=0.95\linewidth]{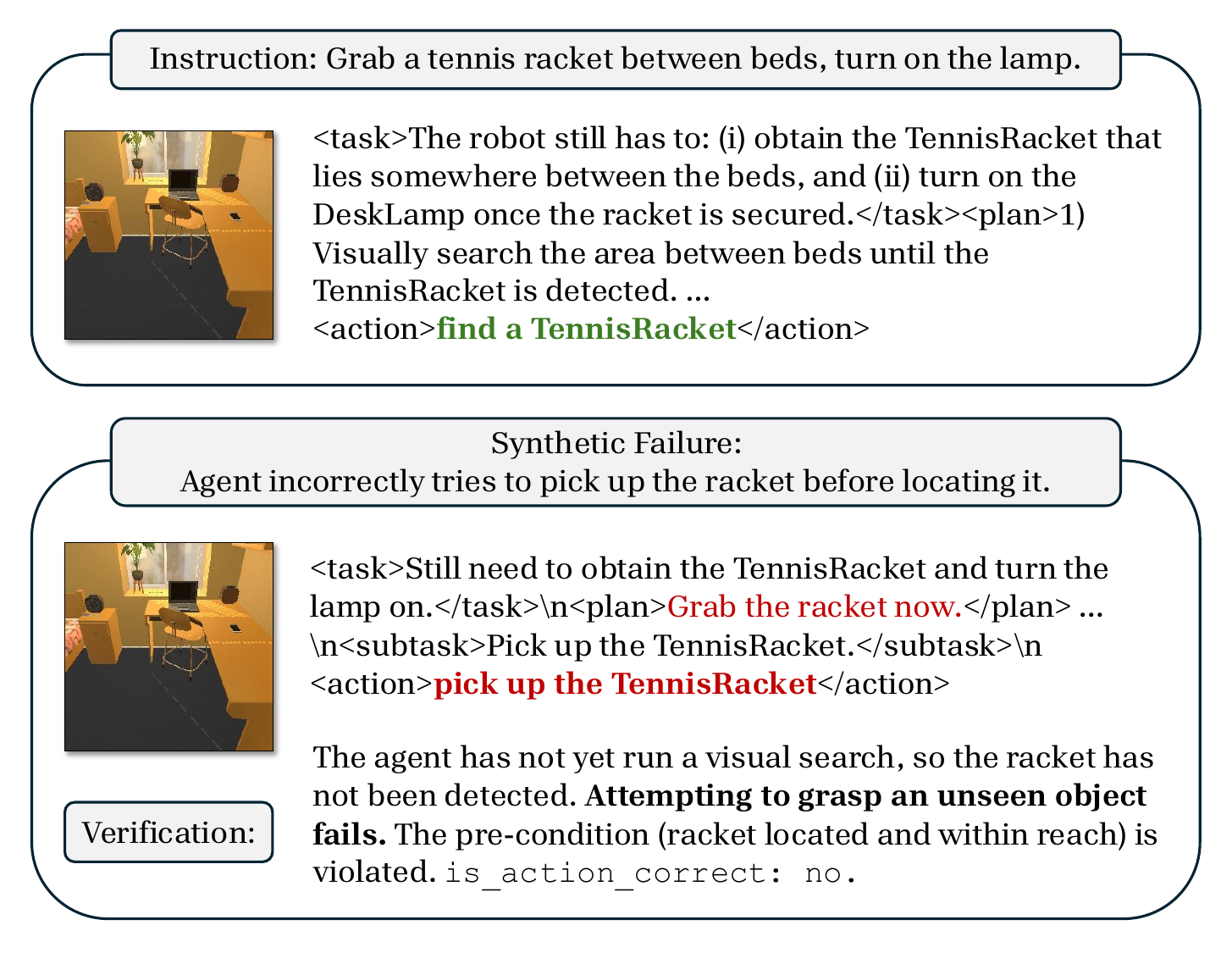}
    \caption{\textbf{Example of a synthetic mistake and verification generated using our pipeline on the ALFRED training dataset.} Starting from a correct action (top; `find a TennisRacket'), our method introduces a mistake (bottom) where the agent does not locate the racket before attempting to pick it. Our method also generates a corresponding verification explaining the mistake.}
    \label{fig:synthetic_mistake_example}
\end{figure}

\begin{figure*}[t]
    \centering
    \includegraphics[width=\linewidth]{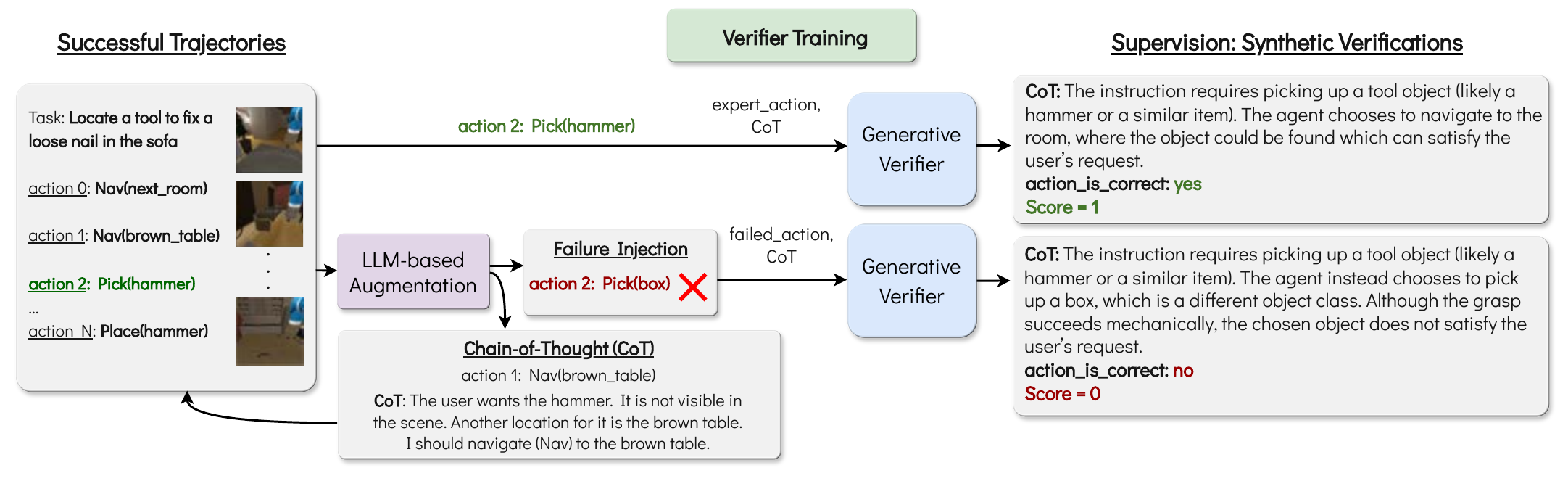}
    \caption{\textbf{Synthetic data generation and training workflow for the verifier.} Successful trajectories are first processed by an LLM to produce chain-of-thought rationales for each action. Then, an LLM introduces realistic and diverse errors into these trajectories and annotates every action with a verification. This dataset is used to train the verifier through supervised finetuning.}
    \label{fig:synthetic_data}
\end{figure*}

The core idea of \ours{} is to augment a base policy with a learned verifier that, at each timestep, evaluates candidate actions and identifies the most reliable one before execution. Because off-the-shelf MLLMs fail as verifiers (as we will show in Sec.~\ref{exp:main_result}), we train a dedicated generative verifier on automatically synthesized failure data (Sec.~\ref{sec:method_synth_data}). Concretely, \ours{} operates via a Best-of-N procedure: at each timestep, the policy samples $N$ candidate actions, the generative verifier~\citep{zhang2025generative,ankner2024critique} evaluates each one by producing a verification reasoning trace followed by a correctness judgement, and the highest-scoring action is executed. Unlike discriminative verifiers that directly output a score~\citep{cobbe2021training, yu2024ovm}, generative verifiers think step-by-step before assigning a score, which has been shown to yield stronger performance while also making the scores more interpretable~\citep{zhang2025generative}.

\subsection{Synthetic Reasoning and Verification Data}
\label{sec:method_synth_data}

\textbf{CoT Augmentation.}
We start with a dataset of successful (`+') trajectories, $\mathcal{D}^+ = \{\tau^+\}$, where a trajectory $\tau^+$ consists of the instruction $I$, and interleaved observations $o$ and actions $a$, $\tau^+ = \{I, o_1, a^+_1, o_2, a^+_2, ... \}$. A model trained on these trajectories will directly output the next action given the observation. However, research in language reasoning and embodied AI has shown that thinking step-by-step can significantly improve the reasoning abilities of models~\citep{wei2022chain,mu2023embodiedgpt}. To train embodied agents that can think step-by-step, similarly to~\citet{zawalski2024robotic}, we prompt a teacher LLM (e.g., OpenAI o3) to augment every action with a chain-of-thought reasoning, $c^+_i$, explaining why the agent should perform the expected action $a^+_i$ given the previous inputs $I, o_1, a^+_1, ..., o_i$.
This gives us a new dataset $\mathcal{D}^+_{CoT} = \{\tau^+_{CoT}\}$, with $\tau^+_{CoT} = \{I, o_1, (c^+_1, a^+_1), o_2, (c^+_2, a^+_2), ... \}$. Note that this procedure only augments every action $a^+_i$ with a chain-of-thought, and does not change the sequence of actions in the trajectories. Unlike~\citet{zawalski2024robotic}, which grounds reasoning traces in visual features such as object and gripper positions for fine-grained manipulation, we target high-level semantic reasoning for tasks requiring long-horizon planning and linguistic interpretation, yielding $\mathcal{D}^+_{CoT}$ (Prompt in Appendix~\ref{prompt:cot_data_gen}).

\textbf{Synthetic Failures for Verifier Training.}
To train a verifier, we require examples of both correct and incorrect actions. Because existing datasets rarely include failed executions, we introduce an automated and scalable pipeline that synthesizes unsuccessful trajectories. For each successful trajectory \(\tau^{+}\), we prompt a large language model (e.g., OpenAI o3) to produce a corresponding failed trajectory \(\tau^{-}\). The model generates realistic and diverse mistakes that span a broad range of failure modes in challenging scenarios, including \emph{wrong object} (e.g., bringing an apple when the task requires a banana), \emph{wrong receptacle} (e.g., placing an item on the sofa instead of the bed), and \emph{precondition violation} (e.g., attempting to turn on a microwave without opening it first). We provide examples of synthetically generated incorrect actions in Figures \ref{fig:synthetic_mistake_example} and \ref{fig:synthetic_data} and Appendix~\ref{app:incorrect_trajectory_examples}. The exact prompts are available in Appendix~\ref{prompt:failed_trajectory}.

For both $\tau^+$ and $\tau^-$, we prompt the model to annotate every action with a verification consisting of chain-of-thought reasoning and a final binary judgement of the form \texttt{action\_is\_correct: yes/no}. These annotated positive and negative samples provide the supervision used to train the verifier.

\subsection{Verifier Training and Inference}
\label{sec:method_verifier_inference}

We fine-tune an MLLM as a verifier that takes as inputs the instruction $I$, all previous actions $a_1, a_2, ..., a_{t-1}$, and the current observation $o_t$, chain-of-thought $c_t$, and action $a_t$ sampled from the policy. It outputs a verification $v_t$ consisting of a chain-of-thought and a verdict. The verifier is trained via supervised finetuning on the data described in Sec.~\ref{sec:method_synth_data} using the same next-token prediction objective as the policy (Sec.~\ref{sec:background}). This process is illustrated in Figure~\ref{fig:synthetic_data}.

During inference, at time $t$, we sample $N$ candidate actions from the policy: $(c^{(1)}_t, a^{(1)}_t), (c^{(2)}_t, a^{(2)}_t), ..., (c^{(N)}_t, a^{(N)}_t)$. We pass each candidate $(c^{(n)}_t, a^{(n)}_t)$ to the verifier and, following the original GenRM procedure~\citep{zhang2025generative}, sample $M$ verifications per action to reduce variance, each consisting of a verification chain-of-thought and a verdict. The verdict can be mapped to a score (`\texttt{yes}' $\rightarrow$ 1 and `\texttt{no}' $\rightarrow$ 0), giving us $M$ scores per action. We average these scores to obtain a final score for every action, $\sigma_t^{(n)}$. Finally, we select the highest-scoring action (Best-of-N): $a_t = \operatorname{argmax}_{n \in [N]} [\sigma_t^{(n)}]$. The selected action $a_t$ is then executed in the environment, and the process repeats at the next timestep.

\section{Experimental Setup}

\subsection{Benchmark Details}

We evaluate our approach on two embodied AI benchmarks targeting out-of-distribution generalization: LangR~\citep{szot2023large} in the Habitat 2.0 simulator~\citep{szot2021habitat}, and ALFRED~\citep{shridhar2020alfred} in the AI2-THOR simulator~\citep{kolve2017ai2}. In both benchmarks, the agent is placed in previously unseen indoor environments and tasked with completing multi-step household tasks (e.g., rearranging objects, examining items) specified through natural language. At each timestep, the agent receives an egocentric RGB observation and selects a high-level semantic action (e.g., \texttt{navigate(table)}, \texttt{pick(apple)}, \texttt{open(fridge)}), which is executed by the simulator.

\textbf{LangR~\citep{szot2023large}.} This benchmark comprises a diverse set of training tasks featuring multiple paraphrastic variations and interactions with different household objects. The benchmark also includes several out-of-distribution (OOD) tasks designed to evaluate the model's generalization capabilities. These evaluation tasks differ from the training set in their natural language instructions, which vary either through linguistic reformulation (termed \textit{Paraphrastic Robustness}, e.g., ``pick up a banana'' $\rightarrow$ ``pick up a yellow curved fruit'') or through changes in the underlying task structure (termed \textit{Behavioral Generalization}, e.g., ``move an apple and a banana'' $\rightarrow$ ``move an apple, a banana, and a ball''). The evaluation suite comprises 8 tasks with 100 instructions each. Further details are available in Appendix~\ref{app:benchmark_details}.

\textbf{ALFRED~\citep{shridhar2020alfred}.} This benchmark, built upon the AI2-THOR simulator~\citep{kolve2017ai2}, encompasses seven distinct task types (such as \textit{pick and place} and \textit{examine in light}) that involve diverse interactions with objects in household environments. In this work, we employ the EB-ALFRED implementation introduced by~\citet{yang2025embodiedbench}, which reorganizes tasks from the original benchmark into several categories designed to assess different aspects of OOD generalization, including \textit{long-horizon tasks}, \textit{common-sense reasoning}, and \textit{spatial understanding}. The evaluation suite comprises 6 tasks with 50 instructions each. Further details are in Appendix~\ref{app:benchmark_details}.

\subsection{Policy and Verifier}

\textbf{Policy Training.}
LangR does not include expert demonstrations for training. To construct its training dataset $\mathcal{D}^+$, we execute an RL-trained policy from~\citet{szot2023large} on the LangR training split, collecting 10K trajectories, each comprising an instruction, observations, and actions. Trajectories in which the agent failed to complete the task are discarded (fewer than 3\% of the total). For EB-ALFRED, we use the instructions, observations, and actions from the training data provided by the original ALFRED benchmark~\citep{shridhar2020alfred}, which contains approximately 6.5K expert demonstrations. For both benchmarks, we prompt OpenAI's o3 model to augment every action in the expert trajectories with a chain-of-thought (see prompts in Appendix~\ref{prompt:cot_data_gen}), yielding $\mathcal{D}^+_{CoT}$ (as described in Sec.~\ref{sec:method_synth_data}). We fine-tune Qwen2.5-VL-3B-Instruct~\citep{bai2025qwen2.5-vl} on $\mathcal{D}^+_{CoT}$ to obtain the chain-of-thought (CoT) policy. Additional implementation and hyperparameter details are provided in Appendix~\ref{app:training_details}.

\textbf{Verifier Training.}
We begin with approximately 4.5K and 6.5K successful trajectories from LangR and ALFRED, respectively. To generate negative samples, we prompt OpenAI's o3 model to synthesize one failed trajectory corresponding to each successful example, and to provide verification annotations for every action within the failed trajectory (see prompts in Appendix~\ref{prompt:failed_trajectory}). This process yields a corpus of failed trajectories, denoted as $\mathcal{D}^-_{CoT}$. The same model is further used to annotate each action in the successful trajectories with corresponding verifications. We combine the verifications from both successful and failed trajectories, and randomly sample from this pool to construct a balanced dataset containing equal numbers of correct and incorrect samples. To ensure a fair comparison, we use the same base model---Qwen2.5-VL-3B-Instruct~\citep{bai2025qwen2.5-vl}---for both the policy and the verifier, differing only in their training data. Additional implementation details are provided in Appendix~\ref{app:training_details}.

\textbf{Inference.}
We use \texttt{vLLM}~\citep{vllm2023} to perform inference with the policy and verifier. For Habitat 2.0, we run experiments on NVIDIA L40 GPUs. For ALFRED, we perform experiments on NVIDIA A100 80GB GPUs. For the No-CoT and CoT policies, we sample the model responses via greedy decoding. When sampling multiple candidate actions, we sample actions and verifications with a temperature of 0.7. For \ours{}, we sample $N=16$ candidate actions and $M=5$ verifications per action at every timestep. We report results and comparisons against baselines in Sec.~\ref{sec:results}.

\section{Experiments}
\label{sec:results}

\begin{table*}[t]
\renewcommand{\arraystretch}{1.45}
\centering
\resizebox{0.85\textwidth}{!}{%
\begin{tabular}{c|c|cccccccc}
\toprule
\multirow{2}{*}{\textbf{Approach}}                               & \multirow{2}{*}{\textbf{Average}} & \multicolumn{4}{c}{\textbf{Paraphrastic Robustness}}                                                                                                                                  & \multicolumn{4}{c}{\textbf{Behavioral Generalization}}                                                                                                                                                                                \\
\cmidrule(lr){3-6}\cmidrule(lr){7-10}
                                                                 &                                   & \textbf{Rephrasing} & \textbf{Context} & \textbf{\begin{tabular}[c]{@{}c@{}}Irrelevant\\ Text\end{tabular}} & \textbf{\begin{tabular}[c]{@{}c@{}}Referring\\ Expressions\end{tabular}} & \textbf{\begin{tabular}[c]{@{}c@{}}Multiple\\ Rearrange\end{tabular}} & \textbf{\begin{tabular}[c]{@{}c@{}}Novel\\ Objects\end{tabular}} & \textbf{\begin{tabular}[c]{@{}c@{}}Multiple\\ Objects\end{tabular}} & \textbf{Conditional} \\
\midrule
\rowcolor{baselinerow}
\multicolumn{10}{c}{\small\textit{Prior Work}} \\[2pt]
\large LLaRP (LLaMa-7B) \cite{szot2023large}           & \large 46 ± 2                            & \large 92 ± 2              & \large 34 ± 2           & \large 32 ± 2                                                             & \large 26 ± 2                                                                   & \large 47 ± 5                                                                & \large 95 ± 4                                                           & \large 0 ± 1                                                               & \large 39 ± 3               \\
\large SemLang (LLaVA-1.5-7B) \cite{szot2024grounding} & \large 58 ± 1                            & \large 92 ± 1              & \large 46 ± 14          & \large 66 ± 6                                                             & \large 31 ± 3                                                                   & \large 80 ± 6                                                       & \large 97 ± 0                                                           & \large 2 ± 2                                                               & \large 46 ± 4               \\
\midrule
\rowcolor{baselinerow}
\multicolumn{10}{c}{\small\textit{Policy only (Qwen-2.5-VL-3B-Instruct)}} \\[2pt]
\large \quad No-CoT                                                      & \large 58                                & \large 93                  & \large 39               & \large 72                                                                 & \large 48                                                                       & \large 68                                                                    & \large 97                                                               & \large 17                                                                  & \large 28                   \\
\large \quad w/ CoT                                                      & \large 65                                & \large 98                  & \large 50               & \large 85                                                                 & \large 59                                                                       & \large 64                                                                    & \large 97                                                               & \large 25                                                                  & \large 42                   \\
\midrule
\rowcolor{baselinerow}
\multicolumn{10}{c}{\small\textit{w/ CoT policy + Verifier (Qwen-2.5-VL-3B-Instruct)}} \\[2pt]
\large \quad + ZS Verifier                                               & \large 64 ± 2                                  & \large 98 ± 1                    & \large 50 ± 3                  & \large 85 ± 5                                                                    & \large 48 ± 4                                                                         & \large 65 ± 4                                                                      & \large 97 ± 0                                                                 & \large 30 ± 4                                                                    & \large 40 ± 3                     \\
\rowcolor{oursrow}
\large \quad \textbf{+ FT Verifier (\ours{})}                             & \large \textbf{71 ± 1}                   & \large \textbf{99 ± 1}     & \large \textbf{52 ± 2}  & \large \textbf{92 ± 1}                                                    & \large \textbf{62 ± 3}                                                          & \large \textbf{82 ± 1}                                                                & \large \textbf{97 ± 0}                                                  & \large \textbf{34 ± 2}                                                     & \large \textbf{48 ± 2}      \\
\bottomrule
\end{tabular}
}
\caption{\textbf{Success rates on the LangR~\citep{szot2023large} benchmark.} The No-CoT, w/ CoT, and \ours{} models are based on Qwen-2.5-VL-3B-Instruct. ZS and FT refer to zero-shot and finetuning, respectively. For verifier-based approaches, results are averaged over three runs. The No-CoT and CoT variants use greedy decoding for action selection, yielding deterministic outcomes. Our proposed approach, which combines chain-of-thought reasoning with a finetuned verifier (\ours{}), consistently outperforms all baselines.}
\label{tab:main_table}
\end{table*}

First, we evaluate the impact of zero-shot and finetuned verifiers on out-of-distribution generalization (Sec.~\ref{exp:main_result}). Next, we examine whether our finetuned verifier can improve larger, off-the-shelf policies it was never trained with (Sec.~\ref{exp:large_policy}). Finally, we present ablation studies examining training pipeline choices, test-time compute scaling, and latency (Sec.~\ref{exp:ablations}).

\subsection{Verifiers Improve Generalization, But Only When Finetuned}
\label{exp:main_result}

We analyze the effect of verification on the LangR benchmark (Table~\ref{tab:main_table}). Fine-tuning with CoT supervision achieves 65\% average success rate, surpassing prior state-of-the-art SemLang~\citep{szot2024grounding} and establishing a strong baseline.

\begin{figure*}[t]
\centering
\footnotesize
\begin{minipage}[t]{0.104\textwidth}
    \centering
    \includegraphics[width=\textwidth]{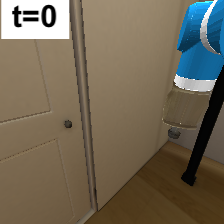}

    \includegraphics[width=\textwidth]{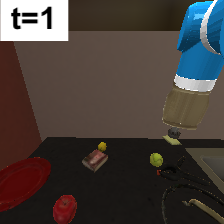}
\end{minipage}%
\hfill
\begin{minipage}[t]{0.8\textwidth}
\fbox{%
\parbox{\dimexpr\textwidth-2\fboxsep-2\fboxrule\relax}{
\vspace{0.1cm}
\textbf{Instruction:} Acquire a \textcolor{blue}{\underline{sports object}} and position it in the brown table.
\vspace{0.2cm}
\hrule
\vspace{0.2cm}
\textbf{Sampled Action:} \texttt{<task>}Acquire a sports object and position it on the brown table; remaining: locate the object, pick it up, navigate to the brown table, and place it.\texttt{</task> <plan>}[search sink, search Table1, search TV stand, search sofa, pick object, navigate to brown table, place object]\texttt{</plan> <subtask\_reason>}Having arrived at the sink, I must check if the object is here and grasp it if present; hence I need to do \textcolor{red}{pick\_sponge(robot\_0)}.\texttt{</subtask\_reason> <subtask>}attempt pick at sink\texttt{</subtask> <action>\textcolor{red}{('pick\_sponge(robot\_0)', 28)}</action>}
\vspace{0.1cm}
\hrule
\vspace{0.1cm}
\textbf{Verification CoT:} The instruction requires picking up a \textcolor{blue}{SPORTS object (likely a ball)}. The chosen action is pick\_sponge, which grasps a sponge instead. It \textcolor{blue}{does not move the correct object} toward completion of the task and therefore fails to advance the goal. \textbf{\texttt{action\_is\_correct: no}}
\vspace{0.1cm}
}%
}
\end{minipage}
\caption{\textbf{Verification example on the LangR benchmark.} Here, the task is to move a ``sports object'', but the agent proposes `\texttt{pick\_sponge}'. The verifier correctly identifies that the sponge is not a sports object, classifying the action as incorrect.}
\label{fig:qualitative_example}
\end{figure*}

We then evaluate whether verification can further improve the CoT policy. Using the same Qwen2.5-VL-3B-Instruct model as a zero-shot verifier (+ ZS Verifier) does not meaningfully improve over the CoT baseline and, in fact, slightly hurts average performance (64\% vs.\ 65\%). This shows that the Best-of-N selection paradigm alone is insufficient; without task-specific training, the verifier cannot reliably distinguish correct actions from incorrect ones. In contrast, equipping the CoT policy with our finetuned verifier (\ours{}) raises performance to 71\%, with consistent gains across all task categories. The improvements are particularly pronounced in challenging scenarios such as \textit{Multiple Objects}, where \ours{} provides roughly a 36\% relative improvement over CoT alone and doubles performance compared to No-CoT. These results demonstrate that the gains of \ours{} stem not from sampling multiple candidates per se, but from our synthetic failure generation and verifier training pipeline. An example in Figure~\ref{fig:qualitative_example} highlights the verifier's effectiveness, where it correctly flags an action arising from the agent misunderstanding the instruction.

\begin{figure*}[t]
\centering
\footnotesize
\begin{minipage}[t]{0.104\textwidth}
    \centering
    \includegraphics[width=\textwidth]{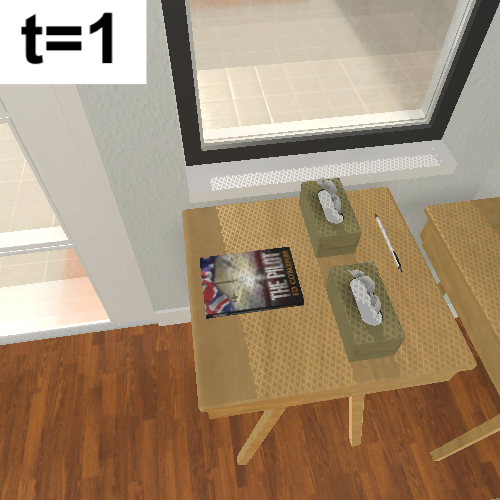}

    \includegraphics[width=\textwidth]{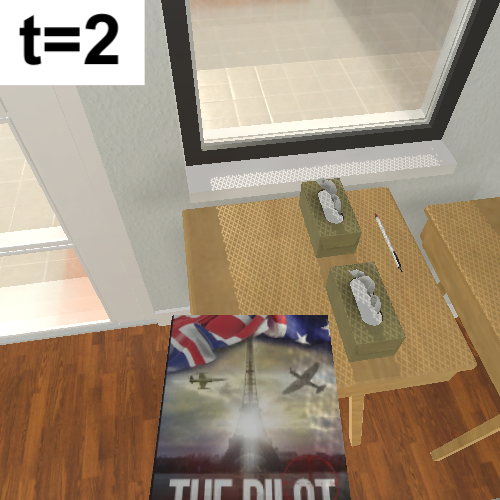}
\end{minipage}%
\hfill
\begin{minipage}[t]{0.8\textwidth}
\fbox{%
\parbox{\dimexpr\textwidth-2\fboxsep-2\fboxrule\relax}{
\vspace{0.1cm}
\textbf{Instruction:} Move the book from the table to the \textcolor{blue}{\underline{large}} comfortable lounging spot.
\vspace{0.2cm}
\hrule
\vspace{0.2cm}
\textbf{Sampled Action:} \texttt{<task>}Book is in hand; it still must be placed on the large comfortable lounge chair.\texttt{</task> <plan>}[1) Find/navigate to the lounge chair, 2) Put the book down]\texttt{</plan> <subtask\_reason>}I cannot place the book until I have reached the correct furniture item. Its exact location relative to the current viewpoint is not confirmed, so I should perform another search targeted at an \textcolor{red}{ArmChair}, hence I should choose the action that finds an \textcolor{red}{ArmChair}.\texttt{</subtask\_reason> <subtask>}Find the lounge chair\texttt{</subtask> <action>}\textcolor{red}{find a ArmChair}\texttt{</action>}.
\vspace{0.1cm}
\hrule
\vspace{0.1cm}
\textbf{Verification CoT:} The instruction specifies the \textcolor{blue}{large} comfortable lounging spot, which in the environment corresponds to the \textcolor{blue}{Sofa} object. Searching for an ArmChair indicates the agent has \textcolor{blue}{mis-identified the target receptacle}. While the action executes successfully, it does not advance the true goal. \textbf{\texttt{action\_is\_correct: no}}
\vspace{0.1cm}
}%
}
\end{minipage}
\caption{\textbf{Verification example on the ALFRED benchmark.} The task is to move the book to a ``large comfortable lounging spot'', but the agent makes a subtle mistake and proposes the related receptacle, armchair. The verifier correctly identifies that the correct receptacle is Sofa, classifying the action as incorrect.}
\label{fig:qualitative_example_alfred}
\end{figure*}

To test whether these findings generalize beyond LangR, we repeat the evaluation on EB-ALFRED (Table~\ref{tab:alfred_main_results}). Our CoT policy, obtained by fine-tuning Qwen2.5-VL-3B-Instruct, achieves an average success rate of 44\%, as shown in Table~\ref{tab:alfred_main_results}. This surpasses Qwen2.5-VL-72B-Instruct (success rate 30\%), a $\sim$20$\times$ larger model, and the best open-weights model reported on EB-ALFRED~\citep{yang2025embodiedbench}, thereby establishing a strong baseline. As on LangR, using the same model as a zero-shot verifier (+ ZS Verifier) does not improve over the CoT baseline (44\% vs.\ 44\%). In contrast, equipping the policy with our finetuned verifier (\ours{}) raises performance to 49\%, confirming that task-specific verifier training is essential for effective verification. To assess whether these findings generalize beyond a single backbone, we repeat the experiment with Gemma-3-4B \citep{gemmateam2025gemma3technicalreport}, fine-tuning both the CoT policy and the verifier. We observe the same trends: the zero-shot verifier provides no meaningful gains, while \ours{} improves the average success rate to 51\%. This consistency across two different model families demonstrates that the benefits of \ours{} are not architecture-specific but arise from the finetuned verifier itself. Taken together, our results on LangR and EB-ALFRED demonstrate that verifying actions at test time can substantially enhance the generalization capabilities of embodied agents in challenging scenarios. Figure~\ref{fig:qualitative_example_alfred} shows an example where the verifier reliably detects a subtle error made by the policy.

\begin{table*}[t]
\renewcommand{\arraystretch}{1.25}
\centering
\resizebox{0.8\linewidth}{!}{%
\begin{tabular}{l|c|cccccc}
\toprule
\textbf{Approach}                  & \textbf{Average}     & \textbf{Base}   & \textbf{\begin{tabular}[c]{@{}c@{}}Common\\ Sense\end{tabular}} & \textbf{\begin{tabular}[c]{@{}c@{}}Complex\\ Instructions\end{tabular}} & \textbf{\begin{tabular}[c]{@{}c@{}}Long\\ Horizon\end{tabular}} & \textbf{Spatial} & \textbf{\begin{tabular}[c]{@{}c@{}}Visual\\ Appearance\end{tabular}} \\
\midrule
\rowcolor{baselinerow}
\multicolumn{8}{c}{\small\textit{Finetuned Policies}} \\[2pt]
Qwen-3B w/ CoT                     & 44                   & 62              & 40                                                              & 58                                                                      & 22                                                              & 34               & \textbf{48}                                                                   \\
+Qwen-3B ZS Verifier                     & 44 ± 2               & 64 ± 2          & 41 ± 7                                                          & 53 ± 5                                                                  & 24 ± 3                                                          & 35 ± 4           & 46 ± 2                                                               \\
\rowcolor{oursrow}
\textbf{+Qwen-3B FT Verifier (\ours{})}              & \textbf{49 ± 2}      & \textbf{67 ± 5} & \textbf{46 ± 2}                                                 & \textbf{62 ± 3}                                                         & \textbf{34 ± 2}                                                 & \textbf{43 ± 3}  & 41 ± 1                                                               \\
\midrule
Gemma-4B w/ CoT                     & 48                 & 62            & 50                                                            & 56                                                                    & \textbf{28}                                                   & 34             & \textbf{56}                                                        \\
+Gemma-4B ZS Verifier                     & 48 ± 2             & 66 ± 0        & 55 ± 1                                                        & 61 ± 5                                                                & 27 ± 3                                                        & 32 ± 4         & 49 ± 6                                                             \\
\rowcolor{oursrow}
\textbf{+Gemma-4B FT Verifier (\ours{})}              & \textbf{51 ± 1}    & \textbf{67 ± 1} & \textbf{56 ± 0}                                               & \textbf{67 ± 1}                                                       & 25 ± 1                                                        & \textbf{37 ± 1} & 53 ± 6                                                             \\
\midrule
\rowcolor{baselinerow}
\multicolumn{8}{c}{\small\textit{Zero-shot Policies}} \\[2pt]
Qwen-2.5-VL-72B            & 30                   & 28              & 38                                                              & 34                                                                      & 12                                                              & \textbf{38}      & 32                                                                   \\
\rowcolor{oursrow}
\textbf{+Qwen-3B FT Verifier (\ours{})} & \textbf{38 ± 1}       & \textbf{45 ± 3}       & \textbf{39 ± 6}                                                & \textbf{44 ± 7}                                                         & \textbf{15 ± 3}                                                 & 36 ± 3                & \textbf{47 ± 1}                                                      \\
% Qwen-2.5-VL-32B            & 13                   & 20              & 10                                                              & 14                                                                      & 8                                                               & 12               & 14                                                                   \\
% \rowcolor{oursrow}
% \textbf{+Qwen-3B FT Verifier (\ours{})} & \textbf{18.67 ± 0.5} & \textbf{24 ± 0} & \textbf{19 ± 7}                                                 & \textbf{20 ± 3}                                                         & 8 ± 6                                                           & \textbf{16 ± 0}  & \textbf{25 ± 4}                                                      \\
Gemma-3-27B                & 19                   & 24              & 24                                                              & 26                                                                      & 0                                                               & \textbf{24}      & 18                                                                   \\
\rowcolor{oursrow}
\textbf{+Qwen-3B FT Verifier (\ours{})} & \textbf{23 ± 0}      & \textbf{30 ± 2} & \textbf{29 ± 1}                                                 & \textbf{27 ± 1}                                                         & \textbf{2 ± 2}                                                  & 23 ± 1           & \textbf{25 ± 1}                                                      \\
InternVL-3.5-38B           & 24                   & 32              & 26                                                              & 26                                                                      & 10                                                              & 18               & 30                                                                   \\
\rowcolor{oursrow}
\textbf{+Qwen-3B FT Verifier (\ours{})} & \textbf{35 ± 2}       & \textbf{38 ± 2}       & \textbf{36 ± 5}                                                & \textbf{39 ± 5}                                                         & \textbf{27 ± 3}                                                 & \textbf{26 ± 2}       & \textbf{41 ± 3}                                                      \\
\bottomrule
\end{tabular}
}
\caption{\textbf{Success rates on EB-ALFRED~\citep{yang2025embodiedbench}.} ZS and FT refer to zero-shot and finetuning, respectively. The CoT policy employs greedy decoding, producing deterministic outcomes. Results involving verifiers are averaged over three runs. Our finetuned verifier consistently outperforms the CoT policy as well as the zero-shot verifier. Further, our finetuned verifier improves performance of larger policies, showing cross-model generalization.}
\label{tab:alfred_main_results}
\end{table*}

% \vspace{-0.4cm}

\noindent\textbf{Verifier-Guided Improvement of Large Policies.}
\label{exp:large_policy}
We evaluate whether a small, finetuned verifier can improve large policies it was never trained with --- a practical setting where large models are inaccessible for fine-tuning. We pair our Qwen2.5-VL-3B-Instruct verifier with several zero-shot policies on EB-ALFRED (Table~\ref{tab:alfred_main_results}). Our verifier consistently improves every policy it is paired with; most notably, it improves Qwen-2.5-VL-72B---a model $\sim$20$\times$ its own size---from 30\% to 38\%, demonstrating that a compact verifier can meaningfully enhance policies far beyond its own scale.

\subsection{Ablation Studies}
\label{exp:ablations}

\begin{figure}[h]
    \centering
    \includegraphics[width=\linewidth]{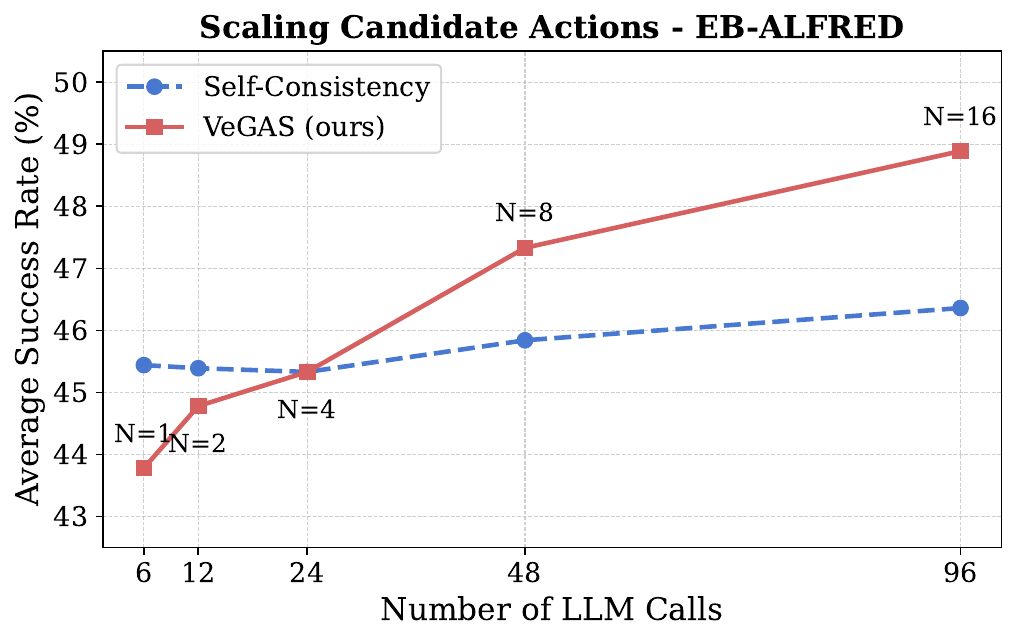}
    \caption{\textbf{Scaling candidate actions on EB-ALFRED.} Average success rate as the number of candidate actions $N$ increases. Both methods use the same total number of LLM calls. \ours{} scales better with compute than Self-Consistency.} % The number of candidate actions $N$ for \ours{} is annotated next to each point.}
    \label{fig:ablation_vary_num_actions}
    % \vspace{-0.4cm}
\end{figure}

\textbf{Impact of Scaling Candidate Actions.}
Increasing the number of sampled candidate actions increases the likelihood of including at least one correct action~\citep{brown2024large}, but also increases inference cost. To isolate whether gains come from the verifier or from sampling diversity alone, we compare against self-consistency~\citep{wang2022self}, which samples multiple actions and selects by majority vote. To ensure a fair comparison, we match total LLM calls across both methods: following~\citet{singhi2025solve}, if \ours{} samples $N$ actions with $M$ verifications each, self-consistency samples $N(M + 1)$ actions.\footnote{The total number of LLM calls for \ours{} is $N + N \times M = N(M+1)$: $N$ policy calls plus $N \times M$ verification calls.} As shown in Figure~\ref{fig:ablation_vary_num_actions}, on EB-ALFRED, self-consistency improves with additional compute but does not scale as efficiently as \ours{}, which exhibits steeper and more consistent gains. These results underscore that a finetuned verifier is crucial for effectively leveraging additional test-time compute to improve embodied agent performance.

\textbf{Sampled Candidates Reliably Contain Correct Actions.}
Best-of-N selection can only succeed if the candidate set contains at least one correct action. We therefore ask: how often does at least one correct action appear among $N$ sampled candidates? Since no ground-truth oracle exists for action correctness, we use o3 as a judge. We measure this coverage probability on LangR as a function of $N$. As shown in Table~\ref{tab:oracle_coverage}, the coverage rises sharply with $N$: with just 10 candidates, at least one correct action is present in 89\% of cases. This confirms that the policy's candidate set is highly likely to contain a correct action, making Best-of-N an effective strategy.

% \vspace{-0.1cm}
\begin{table}[h]
\centering
\footnotesize
\renewcommand{\arraystretch}{0.9}
\begin{tabular}{cc}
\toprule
\textbf{$N$ (candidates)} & \textbf{P(at least one correct)} \\
\midrule
2  & 0.681 \\
4  & 0.794 \\
10 & 0.894 \\
\bottomrule
\end{tabular}
\caption{\textbf{Candidate set coverage on LangR.} Probability that at least one correct action appears among $N$ candidates.}
\label{tab:oracle_coverage}
\end{table}

\textbf{Teacher Model Sensitivity.}
We investigate whether replacing o3 with the cheaper Qwen-3-VL-8B-thinking as teacher degrades verifier quality. As shown in Table~\ref{tab:teacher_sensitivity}, the weaker-teacher verifier achieves 69\%, compared to 71\% with o3-generated data and 65\% for the CoT baseline. While o3 yields the best performance, a much cheaper teacher still provides meaningful gains, making our pipeline accessible without requiring expensive frontier models.

% \vspace{-0.1cm}
\begin{table}[h]
\centering
\footnotesize
\renewcommand{\arraystretch}{0.9}
\begin{tabular}{lc}
\toprule
\textbf{Approach} & \textbf{Avg.\ Success Rate} \\
\midrule
w/ CoT (no verifier)              & 65 \\
+ FT Verifier (Qwen3-8B teacher)   & 69 \\
+ FT Verifier (o3 teacher) & \textbf{71} \\
\bottomrule
\end{tabular}
\caption{\textbf{Teacher model sensitivity on LangR.} Average success rate when using Qwen-3-VL-8B-thinking vs.\ o3 to generate synthetic verifier training data.}% A weaker teacher still yields meaningful gains over the no-verifier baseline.}
\label{tab:teacher_sensitivity}
\end{table}

\textbf{Latency.}
A natural concern with \ours{} is inference latency: sampling $N$ candidate actions and $M$ verifications per action requires $N(M+1)$ total LLM calls. However, since all candidates and verifications can be sampled in parallel, the wall-clock overhead is far more modest than the raw call count suggests. Table~\ref{tab:latency} reports latency as we scale $N$ at $M=5$. Going from $N=1$ (a single greedy action, 1 LLM call) to $N=8$ (48 LLM calls, 48$\times$ more) increases latency by only 2$\times$ (3s $\to$ 6s). This demonstrates that parallel sampling makes \ours{} practical for deployment even at larger compute budgets.

% \vspace{-0.2cm}
\begin{table}[h]
\centering
\scriptsize
\renewcommand{\arraystretch}{0.8}
\begin{tabular}{ccc}
\toprule
\textbf{Config} & \textbf{LLM Calls} & \textbf{Latency} \\
\midrule
$N=1$           & 1                  & 3s               \\
$N=4,\ M=5$     & 24                 & 5s               \\
$N=8,\ M=5$     & 48                 & 6s               \\
$N=16,\ M=5$    & 96                 & 8s               \\
\bottomrule
\end{tabular}
\caption{\textbf{Latency vs.\ compute budget.} Wall-clock time to sample all actions and verifications for increasing $N$ (candidate actions).}
\label{tab:latency}
\end{table}

\textbf{Impact of Visual Input to the Verifier.}
We ask: if we train a verifier that is text-only (i.e., receiving only the action and its reasoning CoT, without the egocentric image), how much does performance degrade? On LangR, the text-only verifier achieves the same average success rate as the multimodal verifier (71\% vs.\ 71\%). On EB-ALFRED, we observe only a marginal drop (49\% vs.\ 47.5\%). We note that the text-only verifier is not truly ``blind'': the chain-of-thought reasoning trace accompanying each candidate action describes the visual scene in natural language, which may explain why removing the image input does not meaningfully hurt performance. This is consistent with prior work showing that text-based scene descriptions are sufficient for high-level embodied planning~\citep{huang2025esca,yang2025embodiedbench}. We also speculate that current high-level benchmarks lack complex scenarios with occlusions or fine-grained visual distinctions where explicit vision-based verification would be most beneficial.

\section{Conclusion}
We introduced \textbf{Verifier-Guided Action Selection (\ours{})}, a test-time framework that improves the out-of-distribution robustness of embodied agents via an explicit verification step. Using an automated pipeline to synthesize failure trajectories for verifier training, \ours{} achieves consistent gains on LangR and EB-ALFRED, including over significantly larger off-the-shelf policies. Our analyses show that verifier finetuning is essential for reliably leveraging additional test-time compute.

\section*{Acknowledgements}

Nishad Singhi is supported by a LOEWE Start-Professur (LOEWE/4b//519/05.01.002-(0006)/94). Marcus Rohrbach is supported in part by an Alexander von Humboldt Professorship in Multimodal Reliable AI, sponsored by Germany's Federal Ministry for Education and Research.
For compute, we gratefully acknowledge support from the hessian.AI Service Center (funded by the Federal Ministry of Research, Technology and Space, BMFTR, grant no. 16IS22091) and the hessian.AI Innovation Lab (funded by the Hessian Ministry for Digital Strategy and Innovation,
grant no. S-DIW04/0013/003). The work has benefited from the Excellence Cluster “Reasonable AI” by the Deutsche Forschungsgemeinschaft (DFG, German Research Foundation) under Germany's Excellence Strategy – EXC-3057, DFG Emmy Noether Programme (CH 2676/1-1), European Union’s Horizon Europe project “MANiBOT” (Grant No.: 101120823), European Union’s Horizon Europe project “ARISE” (Grant No.: 101135959), BMFTR Project “RIG” (Grant No.: 16ME1001).
% We thank the anonymous reviewers for their valuable feedback.

{
    \small
    \bibliographystyle{ieeenat_fullname}
    \bibliography{main}
}

% WARNING: do not forget to delete the supplementary pages from your submission
\clearpage
\hypersetup{pageanchor=false}
\clearpage
\setcounter{page}{1}
\maketitlesupplementary
First, we provide additional details about the benchmarks (App \ref{app:benchmark_details}) and the training setup (App \ref{app:training_details}). Then, we provide the prompts used for synthetic data generation in App \ref{app:prompts}. Further, qualitative examples of synthetic mistakes and verifications are available in App \ref{app:incorrect_trajectory_examples}. Finally, we provide some qualitative examples of the outputs generated by our verifiers at test time in App \ref{app:verifier_examples}.

\section{Details about Benchmarks}
\label{app:benchmark_details}

\subsection{LangR}

The LangR benchmark~\cite{szot2023large}, built on the Habitat 2.0~\cite{szot2021habitat} simulator, is designed to evaluate the generalization capability of embodied agents in household rearrangement scenarios. Agents receive high-level instructions and must execute tasks that involve manipulating objects (pick, open, place), searching for target items, and performing simple forms of logical reasoning such as conditional operations. The allowed actions are: \texttt{navigate(receptacle), pick(object), place(object), open(receptacle), close(receptacle)}.

The actions are executed only if certain preconditions are satisfied. For example, an object can be picked only if it is within reach. The object categories used are: ball,
clamp, hammer, screwdriver, padlock, scissors, block, drill, spatula, knife, spoon, plate, sponge,
cleanser, plum, pear, peach, apple, lemon, can, box, banana, strawberry, lego, rubriks cube, book,
bowl, cup, fork. The maximum number of steps for an episode is 32.

The benchmark provides a suite of training tasks along with a separate set of held-out test tasks. The test split is constructed to examine multiple aspects of generalization, including previously unseen environments and novel instruction formulations.

Two main forms of generalization are emphasized. The first is \textbf{paraphrastic robustness}, where the agent must interpret varied rephrasings of an instruction that shares the same underlying goal. The second is \textbf{behavioral generalization}, which requires the agent to handle new types of reasoning that do not appear in the training distribution. For example, during training the agent may learn to locate a specified number of object instances, while the multiple rearrangements task in the test set requires discovering all instances of a category without being told the exact count. We describe the tasks below.

\textbf{Paraphrastic Robustness}
\begin{itemize}
    \item \textbf{Instruction Rephrasing:} Same underlying goal expressed using a different wording than in training.

    \item \textbf{Referring Expressions:} Objects are mentioned through descriptive or visual attributes rather than their canonical names
    (e.g., a banana described as a curved yellow fruit).

    \item \textbf{Context:} Objects are referred to within a situational or contextual description
    (e.g., a ball described as a sports object).

    \item \textbf{Irrelevant Instruction Text:} Additional text is included that does not affect the task but may distract the agent.
\end{itemize}

\textbf{Behavioral Generalization}
\begin{itemize}
    \item \textbf{Multiple Rearrangements:} Requires rearranging three objects, although training tasks involve only two.

    \item \textbf{Novel Objects:} Introduces new combinations of instructions and object categories that never co-occur in training.

    \item \textbf{Multiple Objects:} Requires manipulating all instances of an object category.
    The agent must search for and detect every instance, a concept not present in training.

    \item \textbf{Conditional Instructions:} Task outcome depends on whether a specified condition holds
    (e.g., if the fridge is open, move the apple to it; otherwise move the orange, and only the required object).
\end{itemize}

The benchmark also includes a spatial reasoning task that uses instructions such as "place the object to the right of the black table." The action space available to the agent does not provide primitives for lateral (left, right, forward, back) movement, which prevents the agent from acquiring meaningful knowledge of the scene layout. As a result, the task is not compatible with the defined action space, so we exclude it from our evaluation.

\subsection{EB-ALFRED}
The ALFRED \cite{shridhar2020alfred} benchmark is built on top of the AI2THOR simulator \cite{kolve2017ai2}. In this work, we use the EB-ALFRED implementation from \cite{yang2025embodiedbench}, which restructures the original tasks into categories designed to probe different aspects of out-of-distribution generalization. The benchmark includes seven task types: \emph{Pick \& Place}, \emph{Stack \& Place}, \emph{Pick Two \& Place}, \emph{Clean \& Place}, \emph{Heat \& Place}, \emph{Cool \& Place}, and \emph{Examine in Light}. Agents operate using eight possible actions: \texttt{pick up}, \texttt{open}, \texttt{close}, \texttt{turn on}, \texttt{turn off}, \texttt{slice}, \texttt{put down}, and \texttt{find}.

EB-ALFRED tasks are grouped into six subsets, each targeting a distinct skill or reasoning capability:

\begin{itemize}
    \item \textbf{Base:} Evaluates core task solving abilities needed to plan and execute low to medium complexity action sequences.
    \item \textbf{Common Sense:} Measures the use of indirect object references grounded in everyday knowledge (for example, describing a refrigerator as "a receptacle that can keep food fresh for several days") and tests the agent's ability to apply such knowledge during instruction following.
    \item \textbf{Complex Instruction:} Contains longer contexts with both relevant and irrelevant details, assessing an agent's ability to extract the intended instruction.
    \item \textbf{Spatial Awareness:} Refers to objects through spatial relations with other items, testing spatial grounding and relational reasoning.
    \item \textbf{Visual Appearance:} Requires identifying objects based on visual attributes such as color or shape.
    \item \textbf{Long Horizon:} Includes tasks requiring extended action sequences, typically more than 15 steps in EB-ALFRED.
\end{itemize}

To construct the benchmark, \cite{yang2025embodiedbench} used the \emph{valid seen} split of ALFRED. A set of 50 tasks with fewer than 15 steps was first selected, from which the common sense and complex instruction subsets were derived. Another 50 tasks with more than 15 steps formed the long horizon subset. Instances for the visual appearance and spatial awareness subsets were chosen directly from ALFRED based on language references to color, shape, or spatial relations. In total, EB-ALFRED contains 300 test instances, uniformly distributed across the six subsets (50 per subset).

\section{Training Details}
\label{app:training_details}

We train both the policy and verifier using the \texttt{LLaMAFactory} framework \cite{zheng2024llamafactory}. Full finetuning is applied while keeping the vision encoder and projection module fixed. All training runs are conducted on 8$\times$NVIDIA L40 GPUs. Training data is formatted as multi turn dialogues using the \texttt{sharegpt} format provided by \texttt{LLaMAFactory}.

The inputs to the policy and verifier consist of prior images and actions rather than earlier chains of thought. To generate data compatible with this interface, each trajectory from the dataset is decomposed into a set of sub conversations, one for each action step. Consider an original trajectory of the form
\[
I,\; o_1,\; (c_1, a_1),\; o_2,\; (c_2, a_2),\; \dots
\]
where $I$ is the instruction, $o_i$ are observations, $c_i$ are chains of thought, and $a_i$ are actions. The $i$th sub conversation contains the instruction along with all observations and executed actions up to step $i$:
\[
I,\; o_1,\; a_1,\; o_2,\; a_2,\; \dots,\; o_i,\; (c_i, a_i).
\]
All chains of thought except the final one, $c_i$, are removed. During training we compute loss only on the last assistant message of each sub conversation, implemented by setting \texttt{mask\_history} to \texttt{True}. Hyperparameters are provided in Table~\ref{tab:train_hparam}.

\begin{table}[h]
\begin{center}
    \caption{Hyperparameters used for training}
\label{tab:train_hparam}
\begin{tabular}{cc}
\hline
\textbf{Hyperparameter}         & \textbf{Value} \\ \hline
Number of GPUs                  & 8              \\
per\_device\_train\_batch\_size & 2              \\
gradient\_accumulation\_steps   & 4              \\
effective batch size            & 8 * 2 * 4 = 64 \\
learning\_rate                  & 1e-5           \\
num\_train\_epochs              & 3              \\
warmup\_ratio                   & 0.1            \\
bf16                            & True           \\
lr\_scheduler\_type             & cosine         \\ \hline
\end{tabular}
\end{center}
\end{table}

\clearpage
\onecolumn

\section{Prompts}
\label{app:prompts}
\subsection{Prompt for CoT data generation}
\label{prompt:cot_data_gen}

\begin{tcolorbox}[
  breakable,
  width=\textwidth,
  boxsep=2mm,
  left=1mm,
  right=1mm,
  title={Prompt to Generate Synthetic Chain-of-Thought (adapted and modified from \cite{zawalski2024robotic})},
  fonttitle=\bfseries
]
\ttfamily
\begin{lstlisting}[basicstyle=\ttfamily\small, breaklines=true]
You're an expert reinforcement learning researcher. You've trained a policy for controlling a robot with an arm to move around and perform tasks in a household environment.
The robot successfully completed a task specified by the instruction: "{instruction}". For that purpose, the
robot executed a sequence of actions. Consecutive moves that were executed are the following:

insert original trajectory with instruction, images, actions


The attached images show what the robot was seeing at a given time step (corresponding to every <image> tag in the trajectory). You can use the images to understand what the robot was seeing, where it was, what it was holding, whether its actions were successful or not, etc.


## General Information and Instructions

1. possible actions are: pick(object), place_on_recep(receptacle), navigate(receptacle), open_fridge(), close_fridge(), open_cabinet(cabinet).
2. Possible objects are: ball, clamp, hammer, screwdriver, padlock, scissors, block, drill, spatula, knife, spoon, plate, sponse, cleanser, plum, pear, peach, apple, lemon, can, box, banana, strawberry, lego, rubrik's cube, book, bowl, cup, mug, orange, lid, toy airplane, wrench.
3. When exploring, select from possible receptacles: [cabinet, drawer 7, cabinet drawer 6, fridge, chair, black table, brown table, TV stand, sink, right counter, left counter]

## Your objective

I want you to annotate the given trajectory with reasoning. That is, for each step, I need to know not only which action should be chosen,
but importantly what reasoning justifies that action choice. I want you to be descriptive and include all the relevant information available.
The reasoning should include the task to complete, the remaining high-level steps, the high-level movements that should be executed and why they are required and any other relevant justification.


### Begin by describing the task

Start by giving an overview of the task. Make it more comprehensive than the simple instruction. Include the activity, 
the objects the robot interacts with, and their relative locations in the environment. Then, describe 
the high-level movements that were most likely executed, based on the task that was completed and the 
primitive movements that were executed. Also, for each high-level movement write a justification for why it should be executed. 
Write an answer for this part using markdown and natural language. Be descriptive and highlight all the relevant details, but ensure that your description is consistent with the trajectory that was  executed, specified above.

### List the reasonings for each step

Finally, for each step describe the reasoning that allows to determine the correct action. For each step describe the remaining part of the objective, 
the current progress, the objects that are still relevant for determining the plan, and the plan for the next steps, based on the available features. Start the reasoning from a high level and gradually add finer features. I need you to be descriptive and very precise. Ensure that the reasoning is consistent with the task and the executed trajectory. 
Write the answer for this part as a Python-executable dictionary. For every step in the initial trajectory there should be exactly one separate item of the form <step id>:<reasoning>. 
Do not group the answers. The final dictionary should have exactly the same set of integer keys as the assistant messages in the original_trajectory above.
The reasoning should be a single string that describes the reasoning in natural language and includes all the required features.

Each reasoning string should have the following form:
- Describe the full task that remains to be completed (but only describe what remains), and place it inside a tag <task>.
- Describe the complete high-level plan for completing the remaining task (the list of remaining high-level steps), and place it inside a tag <plan>.
- Describe why the chosen high-level step should be executed now, which features of the current environment influence that decision, and how it should be done. Place it within a tag <subtask_reason>. First provide appropriate reasoning, and then the name of the action chosen (and not, "I choose action XYZ because ..."; rather, "I need to do XYZ, hence I should choose this action")
- Describe the high-level step that should be executed now (chosen from the list of high-level steps), and place it inside a tag <subtask>.



## Reminders

Keep in mind that the robot might not know where a certain object is, and it may have to move around the environment to search for it. Searching might be one of your subtasks. 

Also, it's possible that the actions chosen by the robot are sometimes suboptimal, you should take that into account. 
Your output reasoning chains should be from the perspective of the assistant, about how it should reason before picking the next action.
Note that the small pebble like thing near the gripper is part of the gripper itself, not a separate item.
The reasoning should be such that first it states why a certain action should be chosen, and then specify that action ("I need to find XYZ, hence, I should do action A" and not "I should do action A. It will help me find XYZ").
Sometimes, the robot would select non-sensical actions, you don't necessarily have to justify such actions (for example by hallucinating objects that don't exist).
You can use information from all time steps to develop your plan, but your final output should be such that the output at time t doesn't assume information about the future (because the robot will think and act one step at a time).
Also, in the original trajectory, after every assistant action, there is a flag telling whether that action was successful or not. You can use this to inform your reasoning chain output. However, note that during deployment, the agent will not have access to this information directly. Instead, it must use its visual input to decide whether the action succeeded. Hence, in your output reasoning chains, when you say that a previous action was successful/failed, you must say that this is based on the image (you can use the ground truth success status, but don't say it in the output).
In the output reasoning trajectory, it is okay if the robot changes the original plan/subtasks as it obtains more information about the environment, or after failures, similar to how an intelligent agent would adapt as it works toward solving a task. The robot should also be capable of recovering from previous errors. Such changes should be clearly stated in the reasoning string.

## Task summary

Here is a breakdown of what needs to be done:

- Describe the task.
- Describe the high-level movements that were executed, based on the completed task and the listed features.
- Describe the plan for the solution that allowed the robot to complete the task successfully.
- For each step on the trajectory, describe the reasoning that leads to determining the correct action. The reasoning should be descriptive and precise. You should provide exactly one reasoning 
string for each ASSISTANT step on the trajectory specified by original_trajectory.
- Just return the dictionary directly, like:
```python
{{
  ...
}}
```
- At the very end of the response, write a single label FINISHED to indicate that the answer is complete.
\end{lstlisting}
\end{tcolorbox}

\subsection{Prompts for Synthetic Failed Trajectory Generation}
\label{prompt:failed_trajectory}

\begin{tcolorbox}[
  breakable,
  width=\textwidth,
  boxsep=2mm,
  left=1mm,
  right=1mm,
  title={Prompt to Generate Synthetic Mistakes and Verifications},
  fonttitle=\bfseries
]
\ttfamily
\begin{lstlisting}[basicstyle=\ttfamily\small, breaklines=true]
Given the above conversation, your job is to create a new trajectory where the agent makes a mistake. Don't change the instruction, just imagine some mistakes and insert them into the new trajectory. 

Your mistakes should be realistic and plausible, i.e., something that the agent is actually likely to do, and not a random mistake. For example: 
- The agent misunderstands the instruction. it interacts with the incorrect object, or with incorrect receptacles, or doesn't perform the expected operations on the objects/receptacles. Here, try to make the "incorrect" actions still somewhat close to the correct actions, e.g., replacing apple with orange instead of remote, or replacing turn on with turn another related action instead of put down the object. 
- The agent doesn't complete a precondition for the next action, e.g., it does not 'find' the object before interacting with it, or doesn't 'open' the microwave or cabinet before putting the object inside. this one is quite important -- basically, think of ways in which the agent might be incompetent: even if it understood the task correctly, it might fail to perform it properly

- The agent messes up the order in which actions should be done. for example, if it has to clean something, it puts the object in the sink, takes it out, and then turns on the faucet

- The agent gets confused between different instances of the same object

- The agent completes only one part of the instruction and not the full instruction

- The agent doesn't do all the actions required for a subgoal, for example, it cleans an object in the sink, but does not pick it up before putting it somewhere else.


Keep in mind that the original trajectory is successful, so if you insert a mistake there, your action should be different from the action given in the original trajectory, because otherwise that action would be correct.

The output should also be in a similar json format. additionally, for every action in the output, add a judgement on whether the action is correct or incorrect. first, provide detailed reasoning, and then the verdict ("action_is_correct: yes or no"). the verdict and judgement should be a single string and should go in a separate key in the dict called 'verification'. also, it's possible that some actions are correct and others are incorrect in the same conversation. 

Also, per conversation, add a mistake_description field, giving a summary of what the mistake was.
Make sure that in the output, before every assistant message there is a user message.
for example:

<new_trajectory> 
mistake_summary: <insert mistake summary> 

{
"content": "...",
"role": "user",
},
{
"content": "...",
"role": "assistant",
"is_action_successful": true,
"verification": "<insert reasoning here>\n
is_action_correct: X" where X is yes or no, 
},


**IMPORTANT**: 
- Judge every action on whether it advances the task or not. so it is possible that the reasoning for the action is incorrect, but it still advances the task, in that case, the action is correct. 
- The mistakes should start from the task interpretation, planning, and subtask reasoning itself, and not like it understands the task but the subtask/action are not aligned with the reasoning. 
- Keep in mind that the output conversation is from the perspective of the agent, and the agent doesn't know it's making a mistake. Make sure that in the output the agent should not seem to know that its plans and actions are not aligned with the instruction. The agent should think its doing the right thing, so it doesn't even know that it misunderstood the instruction. so there should not be anything like "doing X even though the instruction said Y" or "mistakenly doing X". 
- Try to make a 'realistic' mistake - go for an object that is somewhat related to the actual object mentioned in the instruction. 
- In the verification, try to explain your reasoning in as much detail as possible, writing out which object was targeted and which should have been targeted instead. 
- Try to generate diverse types of mistakes. You don't have to restrict yourself to the mistakes mentioned above, feel free to come up with other realistic failure modes. 
- Keep in mind that the original trajectory is successful, so if you insert a mistake there, your action should be different from the action given in the original trajectory, because otherwise that action would be correct. 
- Simply using a different object/receptacle/action might not be considered a mistake. For example, if a lamp is required, both floorlamp and desklamp are correct. 
- Make sure that every assistant message has all the relevant tags (<task>, <plan>, <subtask>, etc., and especially the <action> tag) 
- In the verification, first try to reason about what the instruction specifies and the action taken by the agent, how the two are related, what the agent should have done, and describe in detail why the action is incorrect. 
- Every verification will be used independently of each other, so every verification should be complete in itself (shouldn't assume existence of previous verifications). 

Double check for these important points before writing out the output conversation. It is crucial that these instructions are followed. So check rigorously and strictly your output, and revise them to follow these instructions if necessary.
\end{lstlisting}
\end{tcolorbox}

\section{Examples of Synthetic Incorrect Actions}
\label{app:incorrect_trajectory_examples}

Here, we show some examples of trajectories with synthetic mistakes generated by the LLM to train the verifier.

\subsection{LangR}

\begin{tcolorbox}[
  breakable,
  width=\textwidth,
  boxsep=2mm,
  left=1mm,
  right=1mm,
  title={Example 1 (LangR): The agent erroneously picks up the bowl even though the instruction specifies a cleaner.},
  fonttitle=\bfseries
]
\ttfamily
\begin{lstlisting}[basicstyle=\ttfamily\small, breaklines=true]
Instruction: Shift a cleanser and place it into the specified black table.


action 1: "<task>Move the bowl over to the black table</task>\n<plan>1. Go to right kitchen counter where the bowl usually sits.\n2. Pick up the bowl.\n3. Navigate to the black table.\n4. Place the bowl on the black table.</plan>\n<subtask>Navigate to the right kitchen counter.</subtask>\n<subtask_reason>The bowl is often left on the right side of the kitchen counter, so I will head there first.</subtask_reason>\n<action>('nav(receptacle_aabb_counter_right_kitchen_counter)', 10)</action>"

verification: "The instruction asks for relocating a cleanser, but the assistant plans to move a bowl instead. Nevertheless, navigating to a kitchen counter could logically be a first step for finding a cleanser as well, so this motion does not directly violate task progress.\naction_is_correct: yes",

=====

action 2: "<task>Move the bowl over to the black table</task>\n<plan>[ Pick bowl --> Go to black table --> Place bowl ]</plan>\n<subtask>Pick up the bowl.</subtask>\n<subtask_reason>I can see the bowl on the counter and will grab it.</subtask_reason>\n<action>('pick_bowl(robot_0)', 42)</action>",

verification: "The picked object is a bowl, not the requested cleanser, so this action does not advance the actual task.\naction_is_correct: no"

<following actions omitted for clarity>
\end{lstlisting}
\end{tcolorbox}

\begin{tcolorbox}[
  breakable,
  width=\textwidth,
  boxsep=2mm,
  left=1mm,
  right=1mm,
  title={Example 2 (LangR): The agent erroneously navigates to Table 1 instead of the TV stand after picking up the object},
  fonttitle=\bfseries
]
\ttfamily
\begin{lstlisting}[basicstyle=\ttfamily\small, breaklines=true]
Instruction: Place both a plum and a plate onto the TV stand.

action 1: "<task>Place a plum and a plate on the TV stand.</task> <plan>['navigate to sink counter', 'pick plum', 'navigate toward TV-stand area', 'place plum', 'search for plate', 'place plate']</plan> <subtask>Navigate to sink counter</subtask> <subtask_reason>The plum is normally kept on the kitchen counter near the sink, so the first step is to move there.</subtask_reason> <action>('nav(receptacle_aabb_sink_kitchen_counter)', 9)</action>",

verification: "The instruction requires obtaining the plum first. Moving from the start position to the sink counter is the expected way to reach the plum, so this movement advances the task.\naction_is_correct: yes",

=========

action 2: "<task>Place a plum and a plate on the TV stand.</task> <plan>['pick plum', 'navigate toward TV-stand area', 'place plum', 'search for plate', 'place plate']</plan> <subtask>Pick plum</subtask> <subtask_reason>The plum is visible on the counter and now within reach, so it should be picked up.</subtask_reason> <action>('pick_plum(robot_0)', 30)</action>",

verification: "Grasping the plum is necessary before transporting it anywhere. This action clearly moves the task forward.\naction_is_correct: yes"

==========

action 3: "<task>Deliver the plum.</task> <plan>['navigate to TV-stand area', 'place plum', 'search for plate', 'place plate']</plan> <subtask>Navigate to TV-stand area</subtask> <subtask_reason>The low wooden table in front of the sofa is identified as the surface directly in front of the television, so I should move there to drop the plum.</subtask_reason><action>('nav(receptacle_aabb_Tbl1_Top1_frl_apartment_table_01)', 6)</action>",

verification: "The instruction specifies the TV stand, not Table 1. Navigating to Table 1 therefore heads to the wrong receptacle and does not directly advance the true goal.\naction_is_correct: no",

<following actions omited for clarity>
\end{lstlisting}
\end{tcolorbox}

\subsection{ALFRED}

\begin{tcolorbox}[
  breakable,
  width=\textwidth,
  boxsep=2mm,
  left=1mm,
  right=1mm,
  title={Example 1 (ALFRED): The agent erroneously picks up the baseball instead of the tennis racket, and turns the lamp off instead of on.},
  fonttitle=\bfseries
]
\ttfamily
\begin{lstlisting}[basicstyle=\ttfamily\small, breaklines=true]
Instruction: Grab a tennis racket between beds, turn on the lamp

action 1: "<task>I need to: (1) collect the elongated sporting object I can see between the two beds, and (2) operate the nearby lamp.</task>\n<plan>1) Focus on the bat-shaped item between the beds and secure it. 2) Locate the desk lamp and toggle its power.</plan>\n<subtask>Look for the sporting item I believe to be a BaseballBat.</subtask>\n<subtask_reason>The long, thin handle and rounded head suggest a bat; I will therefore issue a search for a BaseballBat.</subtask_reason>\n<action>find a BaseballBat</action>",

verification: "The instruction requires the agent to locate a TennisRacket, but the agent mistakenly assumes the object is a BaseballBat and executes a search for that instead.  Although the 'find' command itself executes, it advances the wrong goal.\n is_action_correct: no",


============


action 2: "<task>The sporting item has been detected.  I will now secure it.</task>\n<plan>Approach the object and grasp it.</plan>\n<subtask>Pick up the BaseballBat.</subtask>\n<subtask_reason>With the bat in view, gripping it by the handle completes the first half of the instruction.</subtask_reason>\n<action>pick up the BaseballBat</action>",

verification: "No BaseballBat is present; the elongated object is actually a TennisRacket.  Because the target object does not exist, the simulator would return failure.  Attempting this pick-up neither fulfils nor advances the stated task.\n is_action_correct: no",


============


action 3: "<task>With the sporting item handled, I will now take care of the lamp.</task>\n<plan>Find the DeskLamp, then toggle its power.</plan>\n<subtask>Locate the DeskLamp.</subtask>\n<subtask_reason>The lamp is required next; I must bring it into view before interacting.</subtask_reason>\n<action>find a DeskLamp</action>",

verification: "Searching for the lamp is sensible even though the racket hasn't been obtained yet.  The action does move the agent toward one component of the goal, so the step is reasonable.\n is_action_correct: yes"


============


action 4: "<task>The lamp is now in reach.  I will toggle it.</task>\n<plan>Interact with the lamp's switch.</plan>\n<subtask>Switch the DeskLamp off.</subtask>\n<subtask_reason>The lamp appears lit in my view; turning it off will leave it in the desired off state.</subtask_reason>\n<action>turn off the DeskLamp</action>",

verification: "The instruction explicitly says to turn the lamp ON, not OFF.  The agent does the opposite, so even though the environment executes the switch action this contradicts the task requirement.\n is_action_correct: no",
\end{lstlisting}
\end{tcolorbox}

\begin{tcolorbox}[
  breakable,
  width=\textwidth,
  boxsep=2mm,
  left=1mm,
  right=1mm,
  title={Example 2 (ALFRED): The agent erroneously places the wine bottle instead of the vase on the table.},
  fonttitle=\bfseries
]
\ttfamily
\begin{lstlisting}[basicstyle=\ttfamily\small, breaklines=true]
Instruction: Put a glass vase on a table


action 1: "<task>Instruction outstanding: place the vase on a table.</task>\n<plan>1) Locate a suitable decorative vessel; 2) Pick it up; 3) Bring it to a table; 4) Set it down.</plan>\n<subtask>Begin by locating a nice glass vessel (going to look for a WineBottle that will serve as the vase).</subtask>\n<subtask_reason>The shelves often hold tall glass containers; a wine bottle can double as a vase.</subtask_reason>\n<action>find a WineBottle</action>",

verification: "The instruction explicitly says to put a glass vase (object class Glassbottle/Vase) on a table. The assistant chooses to search for a WineBottle instead. Although both are made of glass, a wine bottle is not the requested vase, so this action does not directly progress toward satisfying the instruction.\naction_is_correct: no",


=============


action 2: "<task>Vessel found; next step is to secure it.</task>\n<plan>Pick the bottle up, then head to a table.</plan>\n<subtask>Grasp the WineBottle.</subtask>\n<subtask_reason>The object is within reach and I need to carry it to the table.</subtask_reason>\n<action>pick up the WineBottle</action>",

verification: "Again, the agent manipulates a WineBottle instead of the required glass vase. Picking it up succeeds mechanically, but the chosen object is incorrect for the task.\naction_is_correct: no",


=============


action 3: "<task>Object secured; now need a table.</task>\n<plan>Navigate to a dining table in the room.</plan>\n<subtask>Find a DiningTable.</subtask>\n<subtask_reason>Any table surface will work as a display spot.</subtask_reason>\n<action>find a DiningTable</action>",

verification: "Navigating toward a table is generally required, but because the agent is holding the wrong object, this search does not really bring it closer to completing the instructed task. Therefore the action, while plausible, does not advance the true goal.\naction_is_correct: no",


=============


action 4: "<task>Table located; time to place the decorative bottle.</task>\n<plan>Release the bottle onto the table surface.</plan>\n<subtask>Put the carried bottle down on the table.</subtask>\n<subtask_reason>The receptacle is in reach and I no longer need to transport the item.</subtask_reason>\n<action>put down the object in hand</action>",

verification: "The assistant places the WineBottle on the table instead of a glass vase. The receptacle is correct (a table), but the object does not match the instruction. Consequently, the high-level task remains unsatisfied.\naction_is_correct: no",
\end{lstlisting}
\end{tcolorbox}

\clearpage

\section{Additional Qualitative Examples of Verification During Inference}
\label{app:verifier_examples}

\subsection{LangR}

\begin{figure*}[h]
\centering
\small
\begin{minipage}[]{0.13\textwidth}
    \centering
    \includegraphics[width=\textwidth]{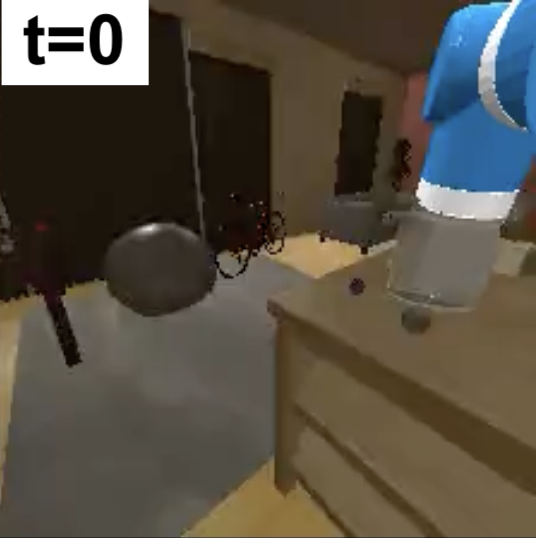}

    \includegraphics[width=\textwidth]{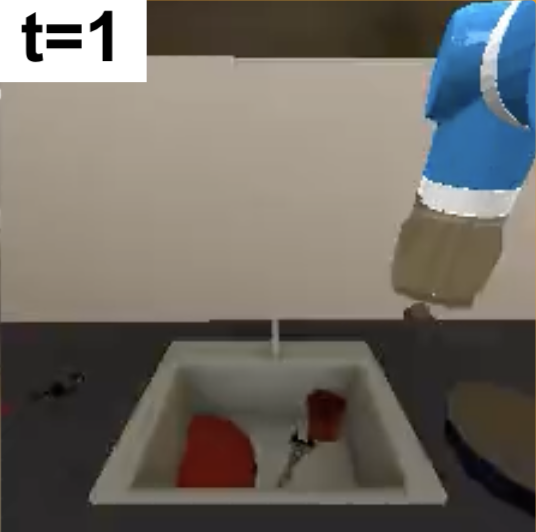}
\end{minipage}%
\hfill
\begin{minipage}[]{0.8\textwidth}
\fbox{%
\parbox{\dimexpr\textwidth-2\fboxsep-2\fboxrule\relax}{
\vspace{0.1cm}
\textbf{Instruction:} Procure a \textcolor{blue}{purple fruit} and deliver it to the left counter.
\vspace{0.2cm}
\hrule
\vspace{0.2cm}
\textbf{Sampled Action:} \texttt{<task>}Find and pick up the purple fruit and then deliver it to the left counter.\texttt{</task> <plan>}[Search TV stand, search sofa, search table2, pick fruit, navigate to left counter, place fruit]\texttt{</plan> <subtask\_reason>}At the TV stand, I should try a pick to see if the fruit is here, hence pick(object).\texttt{</subtask\_reason> <subtask>}Attempt to pick the fruit at the TV stand.\texttt{</subtask> <action>\textcolor{red}{('pick\_strawberry(robot\_0)'}, 38)</action>}
\vspace{0.1cm}
\hrule
\vspace{0.1cm}
\textbf{Verification CoT:} The assistant \textcolor{blue}{tries to pick a strawberry instead of the requested purple fruit}. This does not help fulfil the instruction and therefore does not advance the task.
\textbf{\texttt{action\_is\_correct: no}}
\vspace{0.1cm}
}%
}
\end{minipage}
\caption{Verification example on the LangR benchmark. Here, the task is to move a ``purple fruit'', but the agent tries to pick up a strawberry. The verifier correctly identifies that the strawberry is not the right object, classifying the action as incorrect.}

\end{figure*}

\begin{figure*}[h]
\centering
\small
\begin{minipage}[]{0.13\textwidth}
    \centering
    \includegraphics[width=\textwidth]{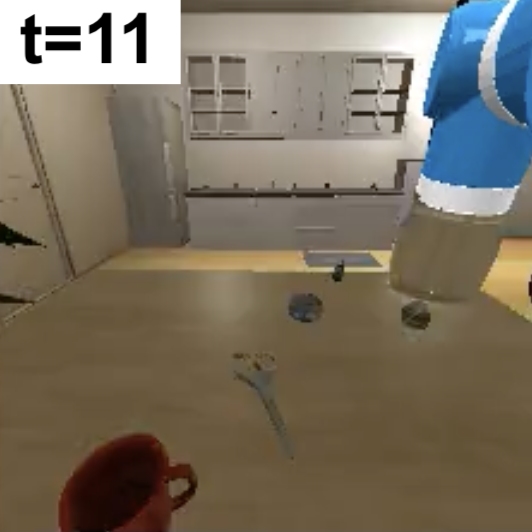}

    \includegraphics[width=\textwidth]{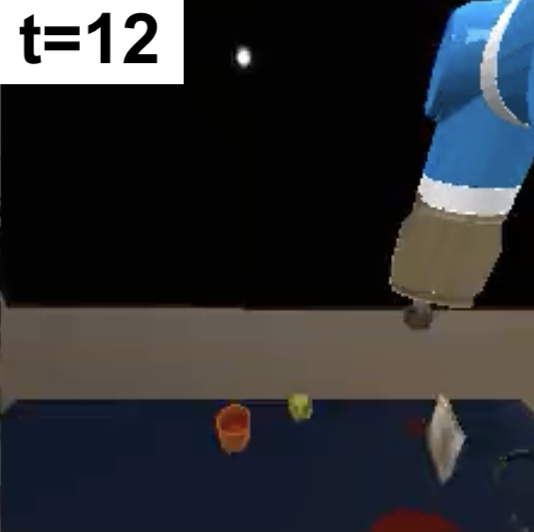}
\end{minipage}%
\hfill
\begin{minipage}[]{0.8\textwidth}
\fbox{%
\parbox{\dimexpr\textwidth-2\fboxsep-2\fboxrule\relax}{
\vspace{0.1cm}
\textbf{Instruction:} Move the rubriks cube to the brown table, the lego to the brown table, and the bowl to the brown table.
\vspace{0.2cm}
\hrule
\vspace{0.2cm}
\textbf{Sampled Action:} \texttt{<task>}Remaining: locate and pick the bowl (it was not found at the TV stand).\texttt{</task> <plan>}[Return to table1 to search for bowl, then cabinet area, pick bowl, navigate to brown table, place bowl]\texttt{</plan> <subtask\_reason>}I am at the brown table and need to check if the bowl is here, hence I should attempt a \textcolor{red}{place action} to see if it is present\texttt{</subtask\_reason> <subtask>}attempt to place bowl on brown table\texttt{</subtask> <action>\textcolor{red}{('place(receptacle\_aabb\_Tbl2\_Top1\_frl\_apartment\_table\_02)'}, 51)</action>}
\vspace{0.1cm}
\hrule
\vspace{0.1cm}
\textbf{Verification CoT:} The robot is not holding any object (the previous pick attempt failed and no new object was picked). \textcolor{blue}{Attempting to place an item while not holding anything cannot succeed and does not progress the task.}
\textbf{\texttt{action\_is\_correct: no}}
\vspace{0.1cm}
}%
}
\end{minipage}
\caption{Verification example on the LangR benchmark. Here, the agent erroneously attempts to place the bowl without successfully picking it up first. The verifier identifies this mistake, classifying the action as incorrect.
}

\end{figure*}

\subsection{ALFRED}

\begin{figure*}[h]
\centering
\small
\begin{minipage}[]{0.13\textwidth}
    \centering
    \includegraphics[width=\textwidth]{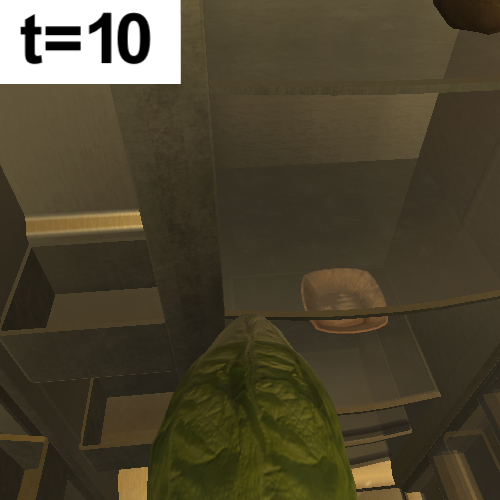}

    \includegraphics[width=\textwidth]{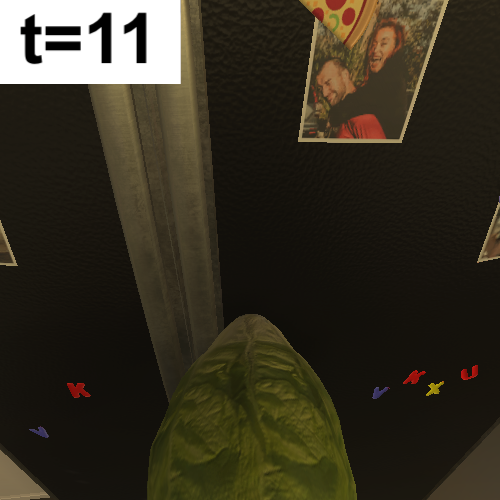}
\end{minipage}%
\hfill
\begin{minipage}[]{0.8\textwidth}
\fbox{%
\parbox{\dimexpr\textwidth-2\fboxsep-2\fboxrule\relax}{
\vspace{0.1cm}
\textbf{Instruction:} \textcolor{blue}{Place} washed leafy green vegetable \textcolor{blue}{in} a receptacle that can keep it fresh for several days
\vspace{0.2cm}
\hrule
\vspace{0.2cm}
\textbf{Sampled Action:} \texttt{<task>}The fridge is open and I (mistakenly) see \textcolor{red}{another lettuce inside}; I still intend to store the held lettuce.\texttt{</task> <plan>}[Optional] Decide whether to store this extra lettuce or switch back to my original intent, then …\texttt{</plan> <subtask\_reason>}I am \textcolor{red}{uncertain of needing both wastebulbs}; \textcolor{red}{closing} keeps the environment tidy while I reconsider.\texttt{</subtask\_reason> <subtask>}Close the fridge door.\texttt{</subtask> <action>}\textcolor{red}{close the Fridge}\texttt{</action>}
\vspace{0.1cm}
\hrule
\vspace{0.1cm}
\textbf{Verification CoT:} \textcolor{blue}{Closing the fridge at this moment is counter-productive because the lettuce needs to remain inside to stay fresh}. Although syntactically valid, it moves the state away from the goal. \textbf{\texttt{action\_is\_correct: no}}
\vspace{0.1cm}
}%
}
\end{minipage}
\caption{Verification example on the ALFRED benchmark. Here, the agent incorrectly attempts to close the fridge before putting the lettuce inside because it mistakenly detects a second lettuce in the fridge. The verifier identifies this mistake, classifying the action as incorrect.
}

\end{figure*}

\begin{figure*}[h]
\centering
\small
\begin{minipage}[]{0.13\textwidth}
    \centering
    \includegraphics[width=\textwidth]{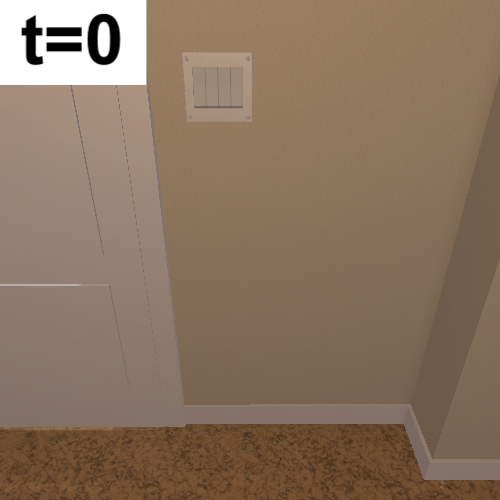}

    \includegraphics[width=\textwidth]{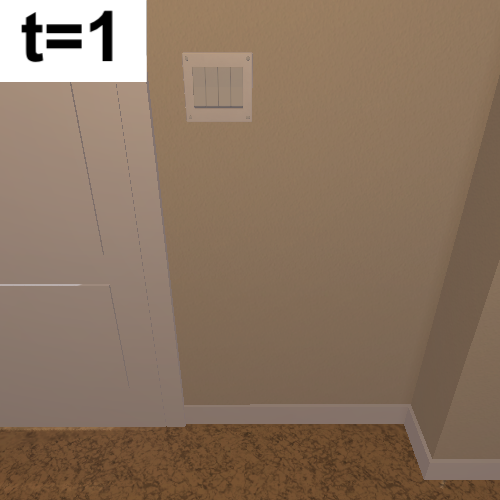}
\end{minipage}%
\hfill
\begin{minipage}[]{0.8\textwidth}
\fbox{%
\parbox{\dimexpr\textwidth-2\fboxsep-2\fboxrule\relax}{
\vspace{0.1cm}
\textbf{Instruction:} As you prepare for dinner, arrange an eating surface that will hold your food with a \textcolor{blue}{spoon} in it on the kitchen table.
\vspace{0.2cm}
\hrule
\vspace{0.2cm}
\textbf{Sampled Action:} \texttt{<task>}Entire task still outstanding: create a state where the dining-table has a container holding a spoon resting on it.\texttt{</task> <plan>}[1] locate a spoon-like utensil, [2] pick it up, [3] find a bowl or cup, [4] place utensil inside it, [5] carry the filled utensil to the dining-table, [6] set it down on the table.\texttt{</plan> <subtask\_reason>}Before anything else I need a spoon; without knowing its position I cannot grasp or use it. Therefore I must visually search for an appropriate utensil first.\texttt{</subtask\_reason> <subtask>}Locate a spoon-like utensil.\texttt{</subtask> <action>}\textcolor{red}{find a Ladle}\texttt{</action>}
\vspace{0.1cm}
\hrule
\vspace{0.1cm}
\textbf{Verification CoT:} The user's instruction explicitly calls for a "\textcolor{blue}{spoon}." The agent \textcolor{blue}{instead decides to search for a Ladle}, which is a different utensil. Although both are eating utensils, \textcolor{blue}{a ladle is not the requested object}. Thus the chosen action does not advance the stated goal directly. \textbf{\texttt{action\_is\_correct: no}}
\vspace{0.1cm}
}%
}
\end{minipage}
\caption{Verification example on the ALFRED benchmark. Here, the task is to move a spoon, but the agent proposes to move a ladle, which is not in the scene. The verifier correctly identifies that the correct object is a spoon, classifying the action as incorrect.}

\end{figure*}

\begin{figure*}[h]
\centering
\small
\begin{minipage}[]{0.13\textwidth}
    \centering
    \includegraphics[width=\textwidth]{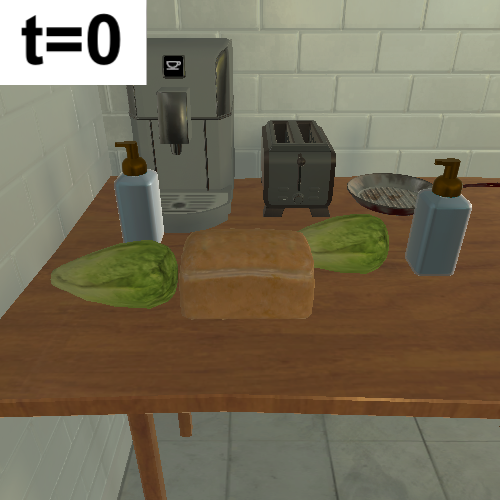}

    \includegraphics[width=\textwidth]{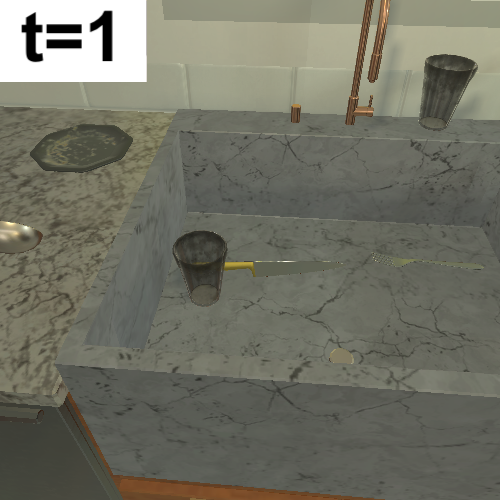}
\end{minipage}%
\hfill
\begin{minipage}[]{0.8\textwidth}
\fbox{%
\parbox{\dimexpr\textwidth-2\fboxsep-2\fboxrule\relax}{
\vspace{0.1cm}
\textbf{Instruction:} During a cooking session, keep a clean \textcolor{blue}{flipping tool} on a wooden table.
\vspace{0.2cm}
\hrule
\vspace{0.2cm}
\textbf{Sampled Action:} \texttt{<task>}Entire task still outstanding: locate a suitable flipping utensil, make sure it is clean, and finally leave it resting on the wooden dining table.\texttt{</task> <plan>}1) \textcolor{red}{Find a knife}. 2) Pick it up. 3) Go to a sink. 4) Wash the knife by turning water on then off. 5) Bring the clean knife to the wooden dining table and set it down there.\texttt{</plan> <subtask\_reason>}The very first requirement is to know \textcolor{red}{where a knife is}; without visual contact I cannot interact with it or progress toward washing or placing. Therefore I need to perform a search action that will move my sensors through the scene until a knife is detected.\texttt{</subtask\_reason> <subtask>}Find a knife\texttt{</subtask> <action>}\textcolor{red}{find a Knife}\texttt{</action>}
\vspace{0.1cm}
\hrule
\vspace{0.1cm}
\textbf{Verification CoT:} The instruction asks for any kind of cleaning flipping tool (\textcolor{red}{i.e., a knife}). By performing the action "find a Knife" the agent begins the search for such an object, which is a reasonable first step toward accomplishing the goal. \textcolor{red}{\textbf{\texttt{action\_is\_correct: yes}}}
\vspace{0.1cm}
}%
}
\end{minipage}
\caption{Verification example on the ALFRED benchmark. Here, the task is to move a spatula, but the agent proposes to move a knife. The \textcolor{red}{verifier fails to identify this mistake} and incorrectly classifies the action as correct.}

\end{figure*}

\end{document}